\documentclass[runningheads]{llncs}
\usepackage{graphicx}

\usepackage{tikz}
\usepackage{comment}
\usepackage{amsmath,amssymb} 
\usepackage{color}
\usepackage{caption}
\usepackage{subcaption}
\usepackage{makecell}
\usepackage{tabularx}
\usepackage{placeins}
\usepackage{mdframed}
\usepackage{nopageno}
\usepackage{siunitx}
\usepackage{xspace}
\usepackage{tabu}
\usepackage{arydshln}
\usepackage{mathtools,xparse}

\usepackage[accsupp]{axessibility}  

\begin{document}
\pagestyle{headings}
\mainmatter
\def\ECCVSubNumber{3658}  

\title{NeFSAC: Neurally Filtered Minimal Samples} 

\titlerunning{NeFSAC: Neurally Filtered Minimal Samples}
%
\author{Luca Cavalli\inst{1} \and
Marc Pollefeys\inst{1,2} \and
Daniel Barath\inst{1}}
\authorrunning{L. Cavalli, M. Pollefeys, D. Barath}
%
\institute{Department of Computer Science, ETH Zurich, Zurich, Switzerland \and
Microsoft Mixed Reality and AI Zurich Lab\\
\email{luca.cavalli@inf.ethz.ch}}

\maketitle

\begin{abstract}

Since RANSAC, a great deal of research has been devoted to improving both its accuracy and run-time.
Still, only a few methods aim at recognizing invalid minimal samples early, before the often expensive model estimation and quality calculation are done. 
To this end, we propose NeFSAC, an efficient algorithm for neural filtering of motion-inconsistent and poorly-conditioned minimal samples. 
We train NeFSAC to predict the probability of a minimal sample leading to an accurate relative pose, only based on the pixel coordinates of the image correspondences. 
Our neural filtering model learns typical motion patterns of samples which lead to unstable poses, and regularities in the possible motions to favour well-conditioned and likely-correct samples. 
The novel lightweight architecture implements the main invariants of minimal samples for pose estimation, and a novel training scheme addresses the problem of extreme class imbalance. 
NeFSAC can be plugged into any existing RANSAC-based pipeline. 
We integrate it into USAC and show that it consistently provides strong speed-ups even under extreme train-test domain gaps -- for example, the model trained for the autonomous driving scenario works on PhotoTourism too. 
We tested NeFSAC on more than 100k image pairs from three publicly available real-world datasets and found that it leads to \textit{one order of magnitude} speed-up, while often finding more accurate results than USAC alone.
The source code is available at \url{https://github.com/cavalli1234/NeFSAC}.



\end{abstract}

\begin{figure}[t]
\centering
\begin{subfigure}{0.3\textwidth}
\centering
\includegraphics[trim={0.3mm 0.3mm 0.3mm 0.3mm},clip,width=\textwidth]{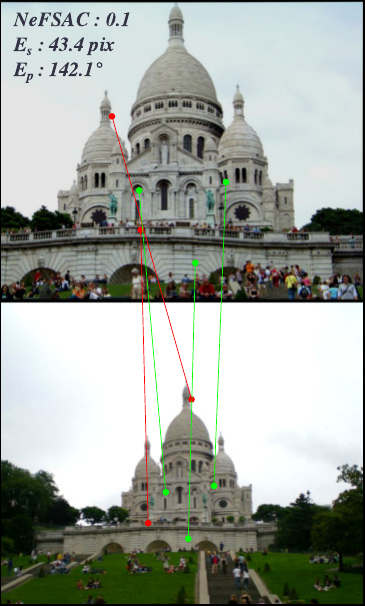}
\caption{With outliers}
\label{fig:teaser_bad}
\end{subfigure}
\begin{subfigure}{0.3\textwidth}
\centering
\includegraphics[trim={0.3mm 0.3mm 0.3mm 0.3mm},clip,width=\textwidth]{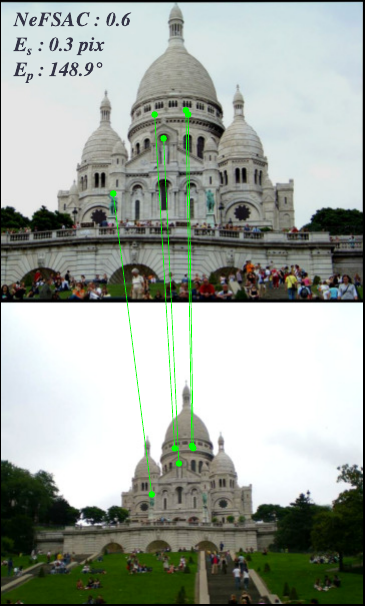}
\caption{Poorly conditioned}
\label{fig:teaser_ill}
\end{subfigure}
\begin{subfigure}{0.3\textwidth}
\centering
\includegraphics[trim={0.3mm 0.3mm 0.3mm 0.3mm},clip,width=\textwidth]{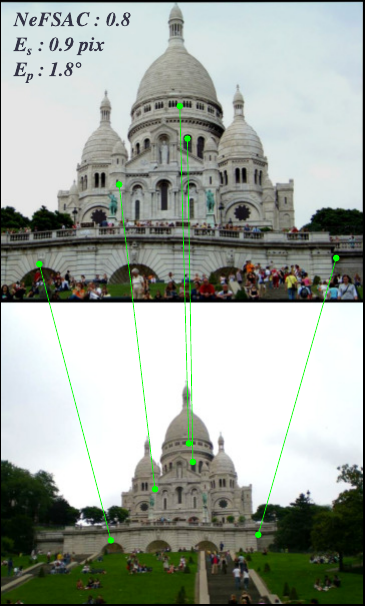}
\caption{Well conditioned}
\label{fig:teaser_good}
\end{subfigure}
\caption{\textbf{Minimal sample filtering}. 
We show three example minimal samples for Essential matrix estimation, with Sampson error $E_s$ in pixels (maximum Sampson error of correspondences with respect to the ground truth Essential matrix), pose error (maximum of rotation and translation error) $E_p$ in degrees, and our model's predicted quality score. \textit{(a)}: a minimal sample encoding unlikely depth and motion due to outliers, easily recognized by NeFSAC only using the pixel's coordinates; 
\textit{(b)}: an all-inlier sample but with two very close correspondences that lead to estimating a poorly conditioned model and, thus, high pose error; \textit{(c)} a minimal sample with widely spaced correspondences. It leads to accurate pose estimation and is strongly preferred by NeFSAC.}
\end{figure}


\section{Introduction}

Robust model estimation is a cardinal problem in Computer Vision. 
RANSAC \cite{fischler1981random} has been a very successful and widely applied approach to robust model estimation since the early days of Computer Vision, and a great research effort has been devoted to improving it. 
While initial efforts \cite{chum2002randomized,chum2005matching,chum2003locally,chum2005two,fischler1981random,matas2005randomized} and some recent works \cite{barath2018graph,barath2021marginalizing,cavalli2020handcrafted,raguram2013usac} are aimed at improving its accuracy, run-time, and robustness attacking well-understood challenges with hand-engineered techniques, more recently we see substantial advancements in augmenting RANSAC for robust model estimation with learning-based techniques \cite{brachmann2019neural,moo2018learning,ranftl2018deep,sarlin2020superglue,zhang2019learning} that derive implicit models directly from data. 
Since one of the biggest challenges in RANSAC is handling large outlier ratios, most of the existing learning-based works are framed as an outlier rejection problem. 
This task is naturally good for learning, since in real scenes correct correspondences are strongly correlated, thus the global set of correspondences can be used to predict which subset is likely to be correct. However, we argue that there is more to be learned from real scenes than recognizing individual outliers.
Another important challenge in RANSAC comes with degenerate and ill-conditioned minimal samples, such as in Figure~\ref{fig:teaser_ill}. 
These configurations often occur in real scenes, e.g., in case of short baseline or when the epipole falls inside the image, when observing close-to-planar scenes or small textured areas leading to localized groups of inlier correspondences.
This is problematic in RANSAC since it means that the often \textit{expensive} model estimation and quality calculation is done unnecessarily on many samples which inherently lead to inaccurate models. 
Also, such models tend to have lots of inliers~\cite{chum2005two,ivashechkin2021vsac}, thus misleading the quality calculation.
%
%
Ideally, the perfect minimal sample filter would be able to recognize and avoid such samples to directly examine well-conditioned ones like the one in Figure~\ref{fig:teaser_good}.

Besides having inherently invalid samples, real-life images tend to follow certain motion patterns that can be also learned and used to further accelerate the robust estimation by rejecting incorrect minimal samples early.
For example, in the autonomous driving scenario when the camera is mounted to a moving vehicle, it follows a distinctive motion pattern that significantly restricts the space of valid minimal sample configurations. 
Even without having such a strong assumption, e.g., when reconstructing internet images, people tend to take pictures approximately aligned with the gravity direction~\cite{ding2021minimal} that, again, gives a probabilistic constraint on the space of valid samples.

Learning a motion prior on minimal samples would allow RANSAC to find the ones that likely lead to the sought model early, spending fewer iterations on unlikely or impossible motions. 
For this purpose, we propose NeFSAC, which learns to filter invalid and motion-inconsistent minimal samples in RANSAC.
NeFSAC can be straightforwardly integrated in any RANSAC variant, e.g. in USAC~\cite{raguram2013usac}, and provide an important speed-up while \textit{improving} accuracy of the estimation. 
We train an extremely lightweight neural network to score minimal samples prior to model estimation, thus being able to screen out thousands of minimal samples with negligible compute time.
In the worst-case scenario, when the domain gap is too huge, NeFSAC degrades back to random filtering, which has no effect on the RANSAC run-time nor on the accuracy. Still, training in new domains requires only the collection of new image pairs with quasi-ground truth poses obtained from any existing pose estimation method.
We integrated our approach into USAC, and measured a reduction of run-time of \textit{one order of magnitude} with significant improvements in the estimation accuracy.

In summary, our contributions are as follows:
(\romannumeral1) We propose NeFSAC, a novel framework to augment RANSAC by learning to efficiently distinguish good minimal samples. Our approach can be seamlessly integrated into any existing RANSAC-based pipeline.
    %
(\romannumeral2) We propose a novel neural architecture for the task, and a novel training scheme for effectively learning the sample quality.
(\romannumeral3) We show that NeFSAC provides impressive speed-ups in RANSAC even without the need for strong motion constraints, while at the same time \textit{improving} the accuracy. In the worst-case scenario, it degenerates to the baseline RANSAC with negligible run-time overhead and no drop in accuracy.

\section{Related Works}

Since RANSAC~\cite{fischler1981random}, great efforts from the research community have been concentrated into improving its components. Many works aim at improving the model scoring technique by modeling inlier and outlier distributions and using likelihood scores instead of the original inlier counting score~\cite{moisan2012automatic,stewart1995minpran,torr2002bayesian,torr2000mlesac}. Similarly, MAGSAC++~\cite{barath2019magsacpp,barath2019magsac,barath2021marginalizing} proposes to marginalize the inlier counting score over a range of possible thresholds, reducing the sensitivity of the scores to the choice of a specific noise scale. LO-RANSAC~\cite{chum2003locally} proposes to perform local optimization of promising models during the search, with later improvements on the cost function and inlier selection~\cite{lebeda2012fixing} and with graph-cut masking of outliers~\cite{barath2018graph}. 
Many of these improvements were combined in USAC~\cite{raguram2013usac} and VSAC~\cite{ivashechkin2021vsac} to achieve state-of-the-art performance.

Closer to our work, another line of research proposes improvements on the sampling scheme to increase the likelihood of detecting an all-inlier sample early.
The most widely used approach is the PROSAC~\cite{chum2005matching} algorithm, where the sampling is biased by prior-established likelihoods (e.g., from ratio-test). 
DSAC~\cite{brachmann2017dsac} first enabled learning through a RANSAC component, followed by Neural-guided RANSAC~\cite{brachmann2019neural} and Deep MAGSAC++~\cite{tong2021deep} which learn inlier sampling likelihoods. 
Other works use spatial techniques to correlate the inlier likelihoods of individual correspondences by preferring neighboring correspondences~\cite{torr2002napsac} or by grouping similar ones~\cite{ni2009groupsac}. 
Such methods combined with early termination techniques~\cite{chum2002randomized,matas2005randomized} can lead to significant improvements in robustness and run-time.
However, none of these works can identify unlikely motions or depth configurations in minimal samples, nor can detect degeneracy of minimal samples. Moreover, we consider our work to be orthogonal to these, since it can be used on top of \textit{any} of these approaches.

The cheirality test~\cite{werner2001cheirality} is widely used to discard some impossible depth configurations. This test discards minimal samples which imply negative depths for some triangulated points. Unfortunately, while it is possible to perform the cheirality test directly on the minimal sample for homographies, it requires expensive epipolar geometry estimation in the cases of Essential matrix or Fundamental matrix estimation.

Even an all-inlier minimal sample can be in a degenerate configuration and lead to unstable relative poses, e.g. when observing a close-to-planar scene.
This problem has been recognized and addressed by DEGENSAC~\cite{chum2005two} that checks for degeneracy and planar configurations of point samples \textit{after} estimating the related epipolar geometry by the plane-and-parallax algorithm~\cite{hartley2003multiple}. Moreover, QDEGSAC~\cite{frahm2006ransac} identifies quasi-degenerate solutions in RANSAC and searches for the missing constraints in the outlier set. Differently from these works, we aim to detect such cases \textit{prior to} the expensive epipolar geometry estimation, providing a consistent speed up to the whole procedure.

Outlier filtering techniques aim to filter the set of putative correspondences to increase the inlier rate prior to robust model estimation. These techniques look for spatial patterns in correspondences and perform explicit spatial verification~\cite{bian2017gms,cavalli2020handcrafted} or learn a spatial verification model~\cite{moo2018learning,zhang2019learning} optionally conditioned on descriptor information to perform matching altogether~\cite{sarlin2020superglue}. Since our approach does not score individual correspondences, but rather joint minimal samples, we consider our contribution to be orthogonal to outlier filtering techniques.

Despite the enormous research devoted to improving robust model estimation with RANSAC, the early selection of minimal samples is still under-explored. Particularly, to the best of our knowledge, no existing work provides a unique solution to embed general motion and depth priors to accelerate RANSAC.
Existing techniques require expensive epipolar geometry estimation to handle degeneracy. In this paper, we show that a lightweight neural network can learn such a filter sufficiently well to provide important savings in run-time \textit{and} improvements in terms of accuracy.

\section{Neural Filtering of Minimal Samples}

In robust model estimation, a set of data points $D = \{ x \in \mathbb{R}^c \}$, optionally contaminated by outliers, is used to fit a model $M \in \mathbb{R}^q$ that minimizes the fitting cost $C = \sum_{x \in D} \mathcal{L}(E(M, x))$ where $E : \mathbb{R}^q \times \mathbb{R}^c \mapsto \mathbb{R}$ is a function that computes the fit error of a data point $x$ with respect to model $M$, and $\mathcal{L} : \mathbb{R} \mapsto \mathbb{R}$ is a robust loss function that generally has small or zero gradients for large errors to minimize the influence of outlier data points. 

Most of the real instances of the robust estimation problem are highly nonlinear, thus the approach proposed by RANSAC~\cite{fischler1981random} is to discretely explore model hypotheses by successively sampling minimal sets of data points $D_{min}^i$ such that they consist of the minimum number $m$ of data points $x$ that can fit exactly a finite set of models. 
The search is then stopped as soon as a satisfactory model has been found according to some termination criterion, and the final model is usually optimized locally to account for all of its inlier data points. 
The reason for using minimal sets stems from the RANSAC termination criterion where the required number of iterations depends exponentially on the sample size to provide probabilistic guarantees of finding the sought model.  

In this work, we aim to drastically reduce the computational expense of such procedure by learning to pre-filter minimal samples before they are used for model estimation or compared to the rest of the data points. 
Notice how this is essentially different from previous works on outlier rejection, that, instead, filter out single data points with the objective of increasing the inlier ratio. 
While our formulation can be applied in general, in the following we will focus on the problems of Essential matrix estimation and Fundamental matrix estimation, where data points are image correspondences ($c=4$), and minimal samples are constituted by, respectively, $m=5$ and $m=7$ correspondences.

In this section, we propose solutions for several challenges that come with the task of learning minimal sample filtering: how to design a lightweight neural architecture that respects all the invariants of minimal samples; how to supervise it in a context of extreme class imbalance; and how to efficiently apply it within RANSAC with the guarantee that, in the worst-case scenario of having a random filter, our method would not cause any degradation of accuracy.

\subsection{Minimal Sample Filtering Network}
\begin{figure}[t]
\centering
\includegraphics[width=0.9\textwidth]{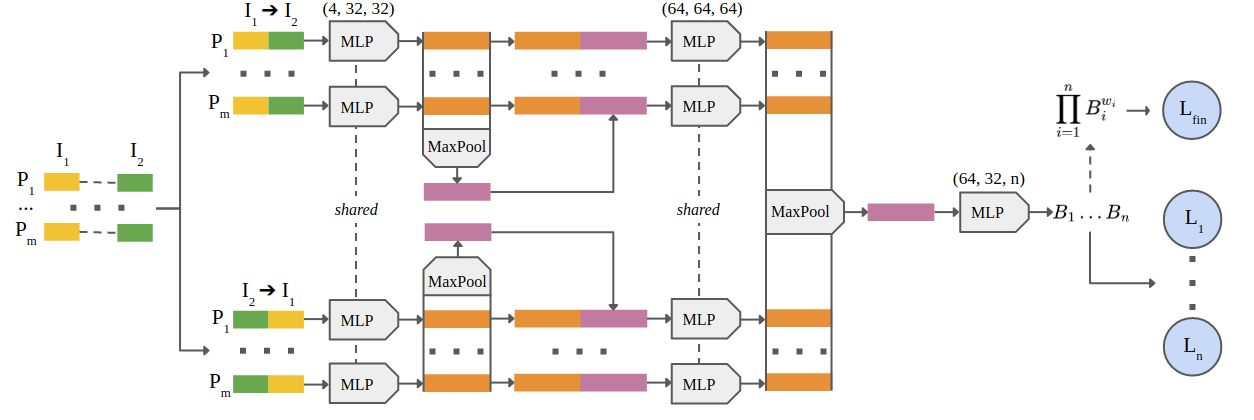}
\caption{\textbf{NeFSAC's network architecture for minimal samples filtering}. We predict the probability of a minimal sample leading to a good pose only taking its coordinates as input. We implement the main invariants of minimal samples with shared MLPs and channel-wise max pooling aggregation. The last MLP outputs $n$ partial scores used during training, whose power-weighted product is the final score for use in RANSAC. The circular nodes represent the binary cross-entropy loss terms with their respective label. We do not propagate gradients across the dashed arrow. }
\label{fig:arch}
\end{figure}

We aim to learn a function $\mathcal{F}: \mathbb{R}^{c \times m} \mapsto [0, 1]$ to score minimal samples, where $c$ is the dimensionality of a data point and $m$ is the minimum number of data points required to fit a finite set of models. Particularly we are interested in Essential matrix estimation ($c=4, m=5$) and Fundamental matrix estimation ($c=4, m=7$). Note that we disregard information about the global configuration of correspondences across the two images: this simplification leads to a faster and smaller model with very little capacity for overfitting. Particularly, since our primary objective is to reduce the computational load in RANSAC, our model of function $\mathcal{F}$ needs to be extremely lightweight. Moreover, our input is structured and the model needs to respect two main invariants: it should be invariant to the ordering of correspondences, and it should be invariant to swapping of the two images. We take inspiration from PointNet~\cite{qi2017pointnet} and frame our main backbone encoder with shared MLPs that embed each correspondence independently in the same feature space, and then correlate them with channel-wise max pooling, thus preserving permutation invariance. Since our second invariant covers only two combinations, we run our backbone encoder with both alternatives and then max pool its features, before going through a final MLP classifier. We keep the network shallow and thin to keep its run-time negligible with respect to the subsequent RANSAC loop. An overall scheme of our network architecture is represented in Figure~\ref{fig:arch}. Note that our final MLP does not output a single score, but several partial scores whose power-weighted product is the final predicted score. This is to contrast the extreme class imbalance that is encountered with random minimal samples with a novel technique that we detail in Section~\ref{sec:supervision}.

\subsection{Data Preprocessing and Network Supervision}
\label{sec:supervision}

We wish to supervise our network to predict a score for each minimal sample, such that higher scores are given to minimal samples which are more likely to lead to a good model estimation. Given a dataset with image correspondences and ground truth poses, the trivial approach would be to solve for the pose of each minimal sample, label them as positive or negative class based on the maximum between rotation and translation angular error, and train a classifier with binary cross-entropy. While extremely simple, we found that in practice this approach suffers from the extreme class imbalance. First, this is due to the fact that with an inlier ratio of $r$ a minimal sample consisting of $m$ data points has an exponentially lower inlier ratio of $r^m$. Second, not every all-inlier minimal sample leads to accurate or even meaningful models, as shown in Figure~\ref{fig:teaser_ill}.
Depending on the demanded accuracy to define a positive sample, this problem can lead to imbalance rates in the order of the \textit{hundreds} in real datasets, making traditional techniques for unbalanced classification insufficient. Our observation aligns well with the common intuition of how many RANSAC iterations are required in practice to ensure a meaningful estimation of relative pose.
We tackle this challenge by proposing to split the prediction of our network into multiple branches: one branch $B_1$ predicts if the minimal sample is constituted of all inliers (labeled on the maximum Sampson error of its correspondences with respect to the ground truth model), and a second branch $B_2$ predicts if the minimal sample leads to a good estimation of the pose, \textit{given that} it is constituted of only inliers. In this setting, the first branch learns to score down minimal samples of impossible or unlikely motions, without suffering from the extra imbalance and complexity coming from ill-conditioned samples. The second branch, trained only on full-inlier samples, learns to score down the ill-conditioned configurations leading to noisy models. We underline the importance of this second branch, since such configurations are not only common in practical scenarios, but even detrimental for RANSAC, since they can collect a large consensus over the image~\cite{chum2005two,ivashechkin2021vsac} and lead to erroneous early termination. For this reason, our approach not only can improve run-time, but it can also improve accuracy and robustness of the RANSAC pipeline it is used into.

In some scenarios, the possible real motions can be partly constrained with expert knowledge which could be useful as a prior to our network. For example, in an autonomous driving context we can \textit{usually} assume that both rotation and translation only happen around the vertical axis. We propose to integrate expert knowledge into the minimal sample filter by the use of additional branches $B_3 \dots B_n$, where each branch is tasked with predicting the adherence of a minimal sample to the analytical model defined by the expert. This extra supervision biases the feature extraction network to find features that can be discriminative also for the expert guidance, thus helping every branch with generalization. 
We detail the expert models used for the autonomous driving application and for PhotoTourism in the supplementary materials.

Finally, since a good sample is composed of all inliers, leads to a good final pose and is conform to expert models, we predict the final score as the product of all the partial scores. Since the different predictive power of each term is not known a-priori, we weight the product of the branches $B_1 \dots B_n$ with weights $w_1 \dots w_n$ at the exponents (i.e., we make a linear combination in log space) and supervise it to predict samples which are both inliers and lead to accurate models. Moreover, we do not propagate gradients to the branch scores $B_1 \dots B_n$ to avoid unstable gradients from the power terms, therefore the branch terms are learned independently of the aggregate score.
Overall, our loss function is:

\begin{equation}
    \sum_i{\mathcal{X}\left(B_i,~ l_i\right)} +  \mathcal{X}\left(\prod_i{B_i^{w_i}}, ~l_1l_2\right)
\end{equation}
\label{eq:losses}

Where $B_i$ are the output branches with respective assigned labels $l_i$ and learned branch weights $w_i$, and $\mathcal{X}$ is the class-weighted cross-entropy loss. Note that, in Equation~\ref{eq:losses}, index $i=1$ refers to the supervision on Sampson error, and index $i=2$ refers to the supervision on pose error which is only applied to branch $B_2$ when the minimal sample is outlier-free. The calculation of assigned labels $l_i$ is detailed in the supplementary materials. We did not experiment with tuning different weights for the losses of every branch.

\subsection{Filtering Minimal Samples in RANSAC}

Since our model learns to score minimal samples by the probability that they will lead to a successful pose estimate, in RANSAC we are interested in exploring high-scoring minimal samples first, and have a termination criterion to stop iterating when an accurate model is found. 
We iteratively take $N$ minimal samples, sort them according to the score predicted by the network, and only process in RANSAC the first $k \ll N$, after which a new batch of $N$ minimal samples is taken only if necessary --  as controlled by the RANSAC termination criterion.
This procedure guarantees that even in the worst case, when the actual motion does not conform with the learned one and the model degrades back to a random filter, RANSAC still finds the sought model eventually. 

We found experimentally that good values are $N=10000$ and $k=500$, leading to aggressive filtering, but much lower values ($N=128$, $k=12$) also work reasonably well for compute-constrained applications. Processing one full batch takes $1.5ms$ on a RTX2080 GPU, or $20ms$ on an i7 7700K CPU. For simplicity we did not experiment adaptive strategies.

The proposed filtering can be straightforwardly combined with the state-of-the-art pre-emptive model verification strategies and samplers.
We use the ones proposed in USAC~\cite{raguram2013usac}, i.e., Sequential Probability Ratio Test~\cite{chum2008optimal}, PROSAC sampling~\cite{chum2005matching} and, also, LO-RANSAC~\cite{chum2003locally} to find accurate results.

\section{Experiments}

In this Section we provide experimental insights into NeFSAC and its impact when integrated into a state-of-the-art RANSAC. We first investigate the quality of its filtering on a pool of random minimal samples, and show that it can improve its average precision (as defined in Section~\ref{sec:ablation}) by over two times in photo collection scenarios (PhotoTourism~\cite{snavely2006photo}) and by over one order of magnitude in strongly motion-constrained scenarios (KITTI~\cite{geiger2012we}). Moreover, the filtering quality generalizes well across extremely different domains. Second, we validate the performance of NeFSAC when integrated in USAC~\cite{raguram2013usac}, and observe \textit{one order of magnitude} speed-up in practice, together with a \textit{significant improvement} in estimation accuracy.

\subsection{Filtering Accuracy and Ablation Study}
\label{sec:ablation}

\begin{figure}[t]
\centering
\begin{subfigure}{0.43\textwidth}
\centering
\includegraphics[trim={5mm 1mm 11mm 13mm},clip,width=\textwidth]{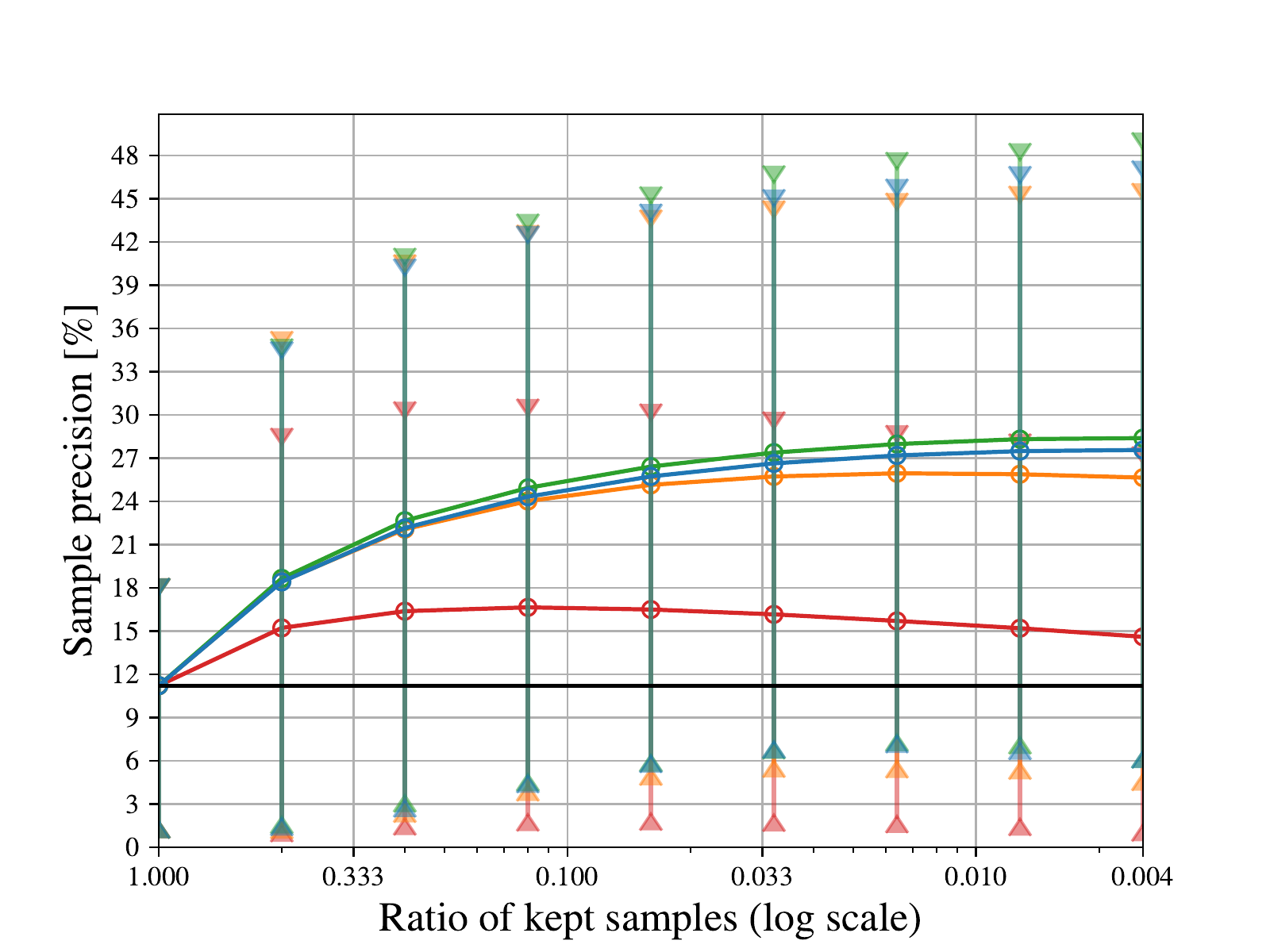}
\caption{Precision on PhotoTourism}
\label{fig:phototourism_filter_accuracy}
\end{subfigure}
\begin{subfigure}{0.43\textwidth}
\centering
\includegraphics[trim={5mm 1mm 11mm 13mm},clip,width=\textwidth]{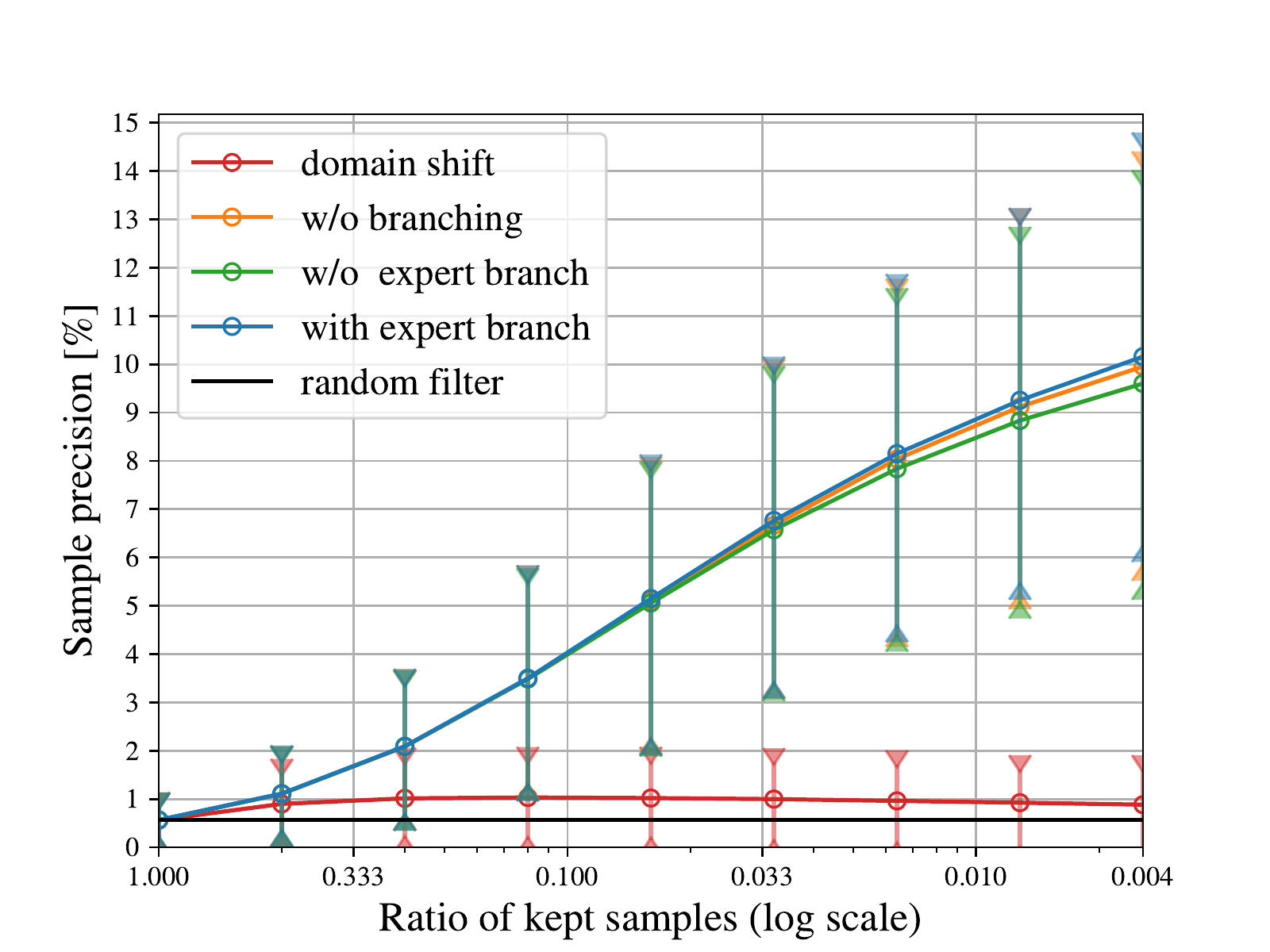}
\caption{Precision on KITTI}
\label{fig:kitti_filter_accuracy}
\end{subfigure}
\caption{\textbf{Precision of neural filtering.} From a pool of minimal samples for E estimation, we keep the highest-scoring minimal samples according to our model and measure the precision of the kept set, i.e. the rate of samples with less than $10^\circ$ of rotation and angular translation error, and less than 2 pixels of Sampson error. The distribution of results across images is represented with solid lines for the average and vertical lines for the two middle quartiles. Our method improves the precision of the minimal sample pool by \textbf{over one order of magnitude} in the autonomous driving scenario, and over two-fold on PhotoTourism.}
\end{figure}

\begin{figure}[t]
\centering
\begin{subfigure}{0.8\textwidth}
\centering
\includegraphics[width=\textwidth]{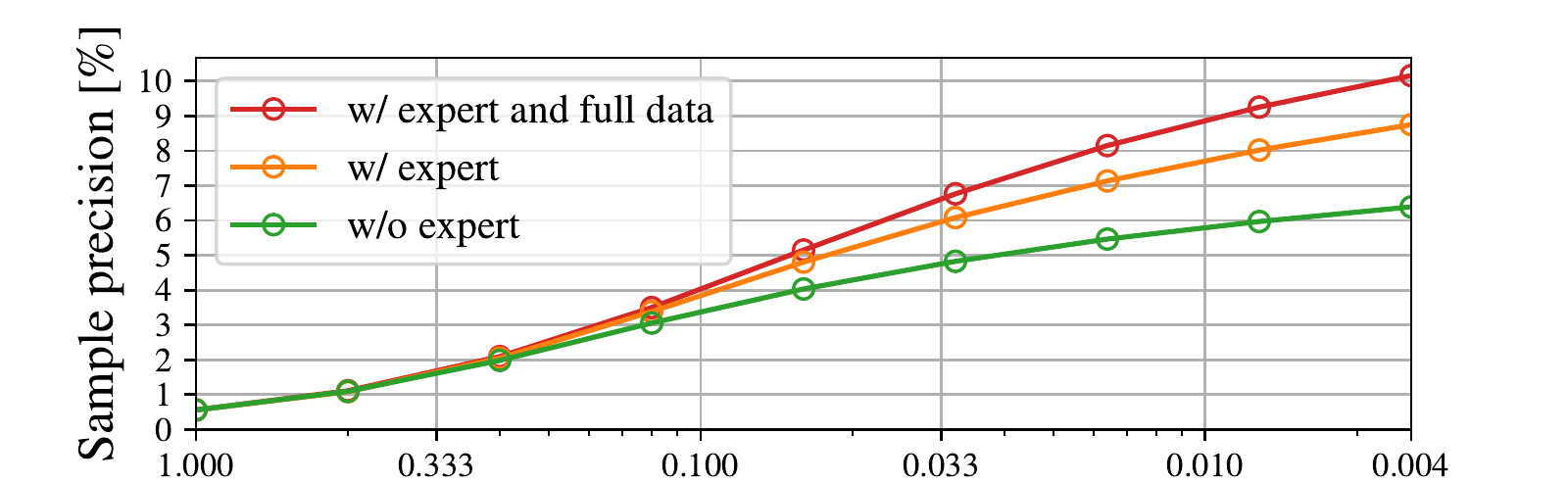}
\caption{Precision on PhotoTourism}
\label{fig:expert_branch_littledata}
\end{subfigure}
\caption{\textbf{Impact of expert branches.} The same evaluation as in Figure~\ref{fig:kitti_filter_accuracy}, performed on KITTI and training with only 250 image pairs from sequence 0 and frame difference 4. The expert branch significantly helps preserving the filtering accuracy in data-scarce conditions. We show only averages for clarity.}
\end{figure}

\begin{figure}[t]
	\begin{center}
	\includegraphics[width=0.405\columnwidth,trim={0mm 0.1mm 0mm 0mm},clip]{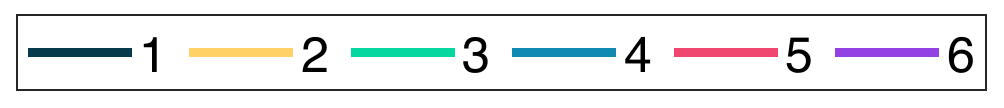}\includegraphics[width=0.39\columnwidth]{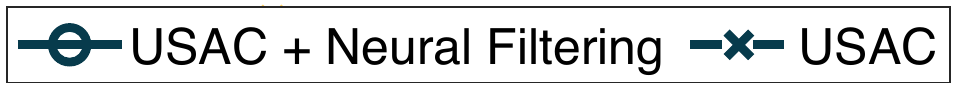}
	\begin{subfigure}[t]{0.93\columnwidth}
		\includegraphics[width=0.32\columnwidth,trim={1mm 0mm 10mm 1mm},clip]{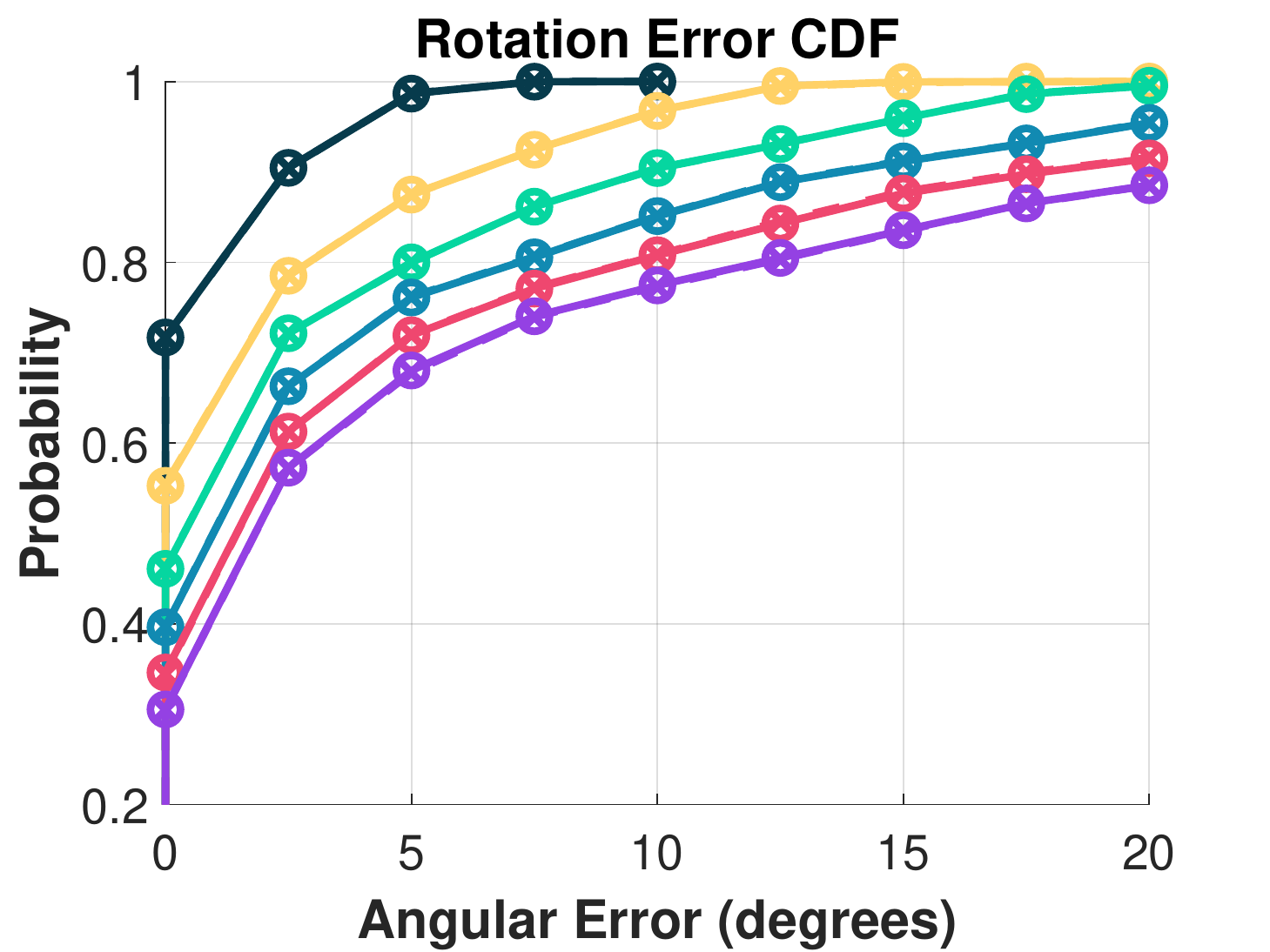}
		\includegraphics[width=0.32\columnwidth,trim={1mm 0mm 10mm 1mm},clip]{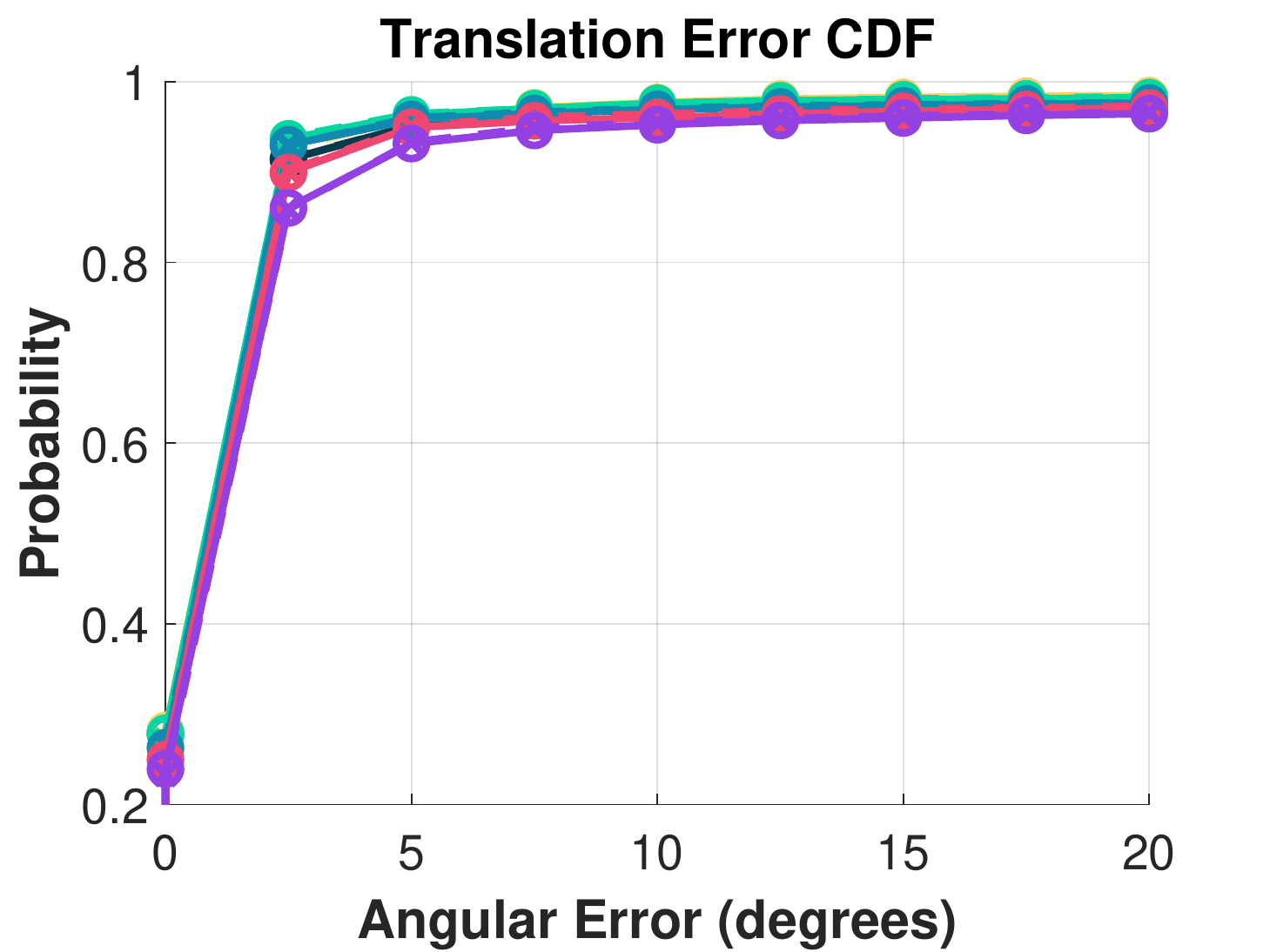}
		\includegraphics[width=0.32\columnwidth,trim={1mm 0mm 10mm 1mm},clip]{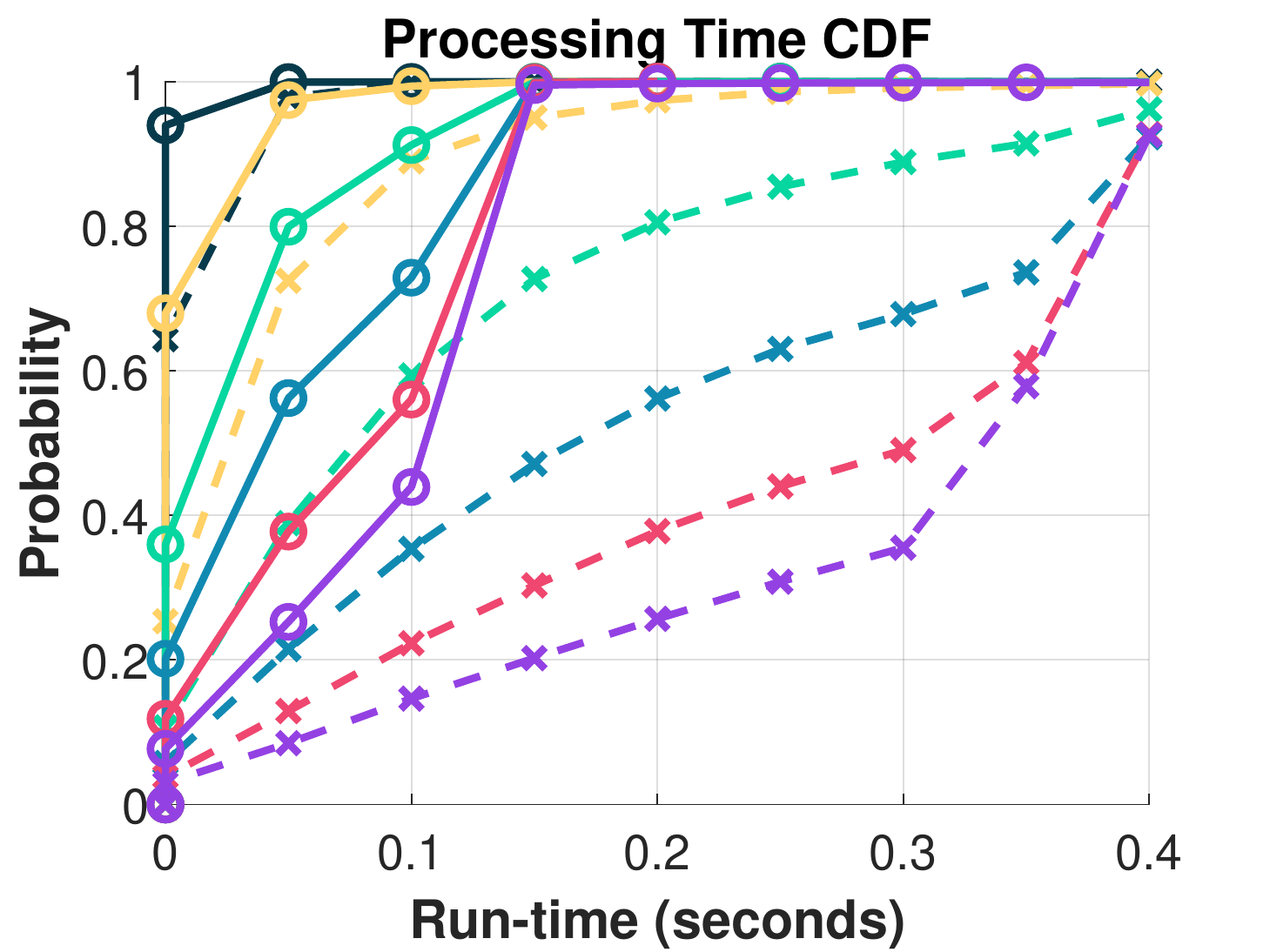}
		\caption{Essential matrix estimation}
	\end{subfigure}
	\begin{subfigure}[t]{0.93\columnwidth}
		\includegraphics[width=0.32\columnwidth,trim={1mm 0mm 10mm 1mm},clip]{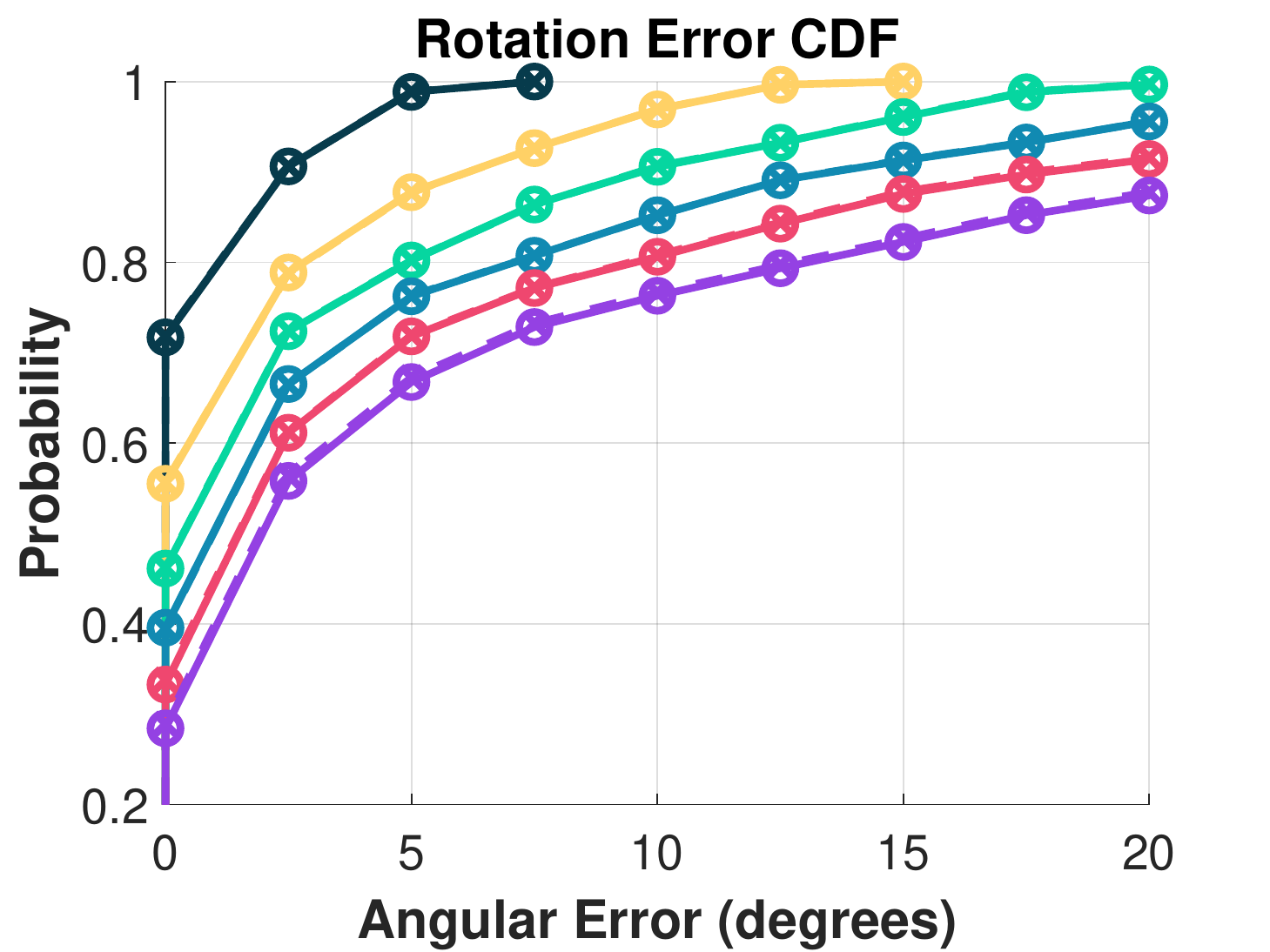}
		\includegraphics[width=0.32\columnwidth,trim={1mm 0mm 10mm 1mm},clip]{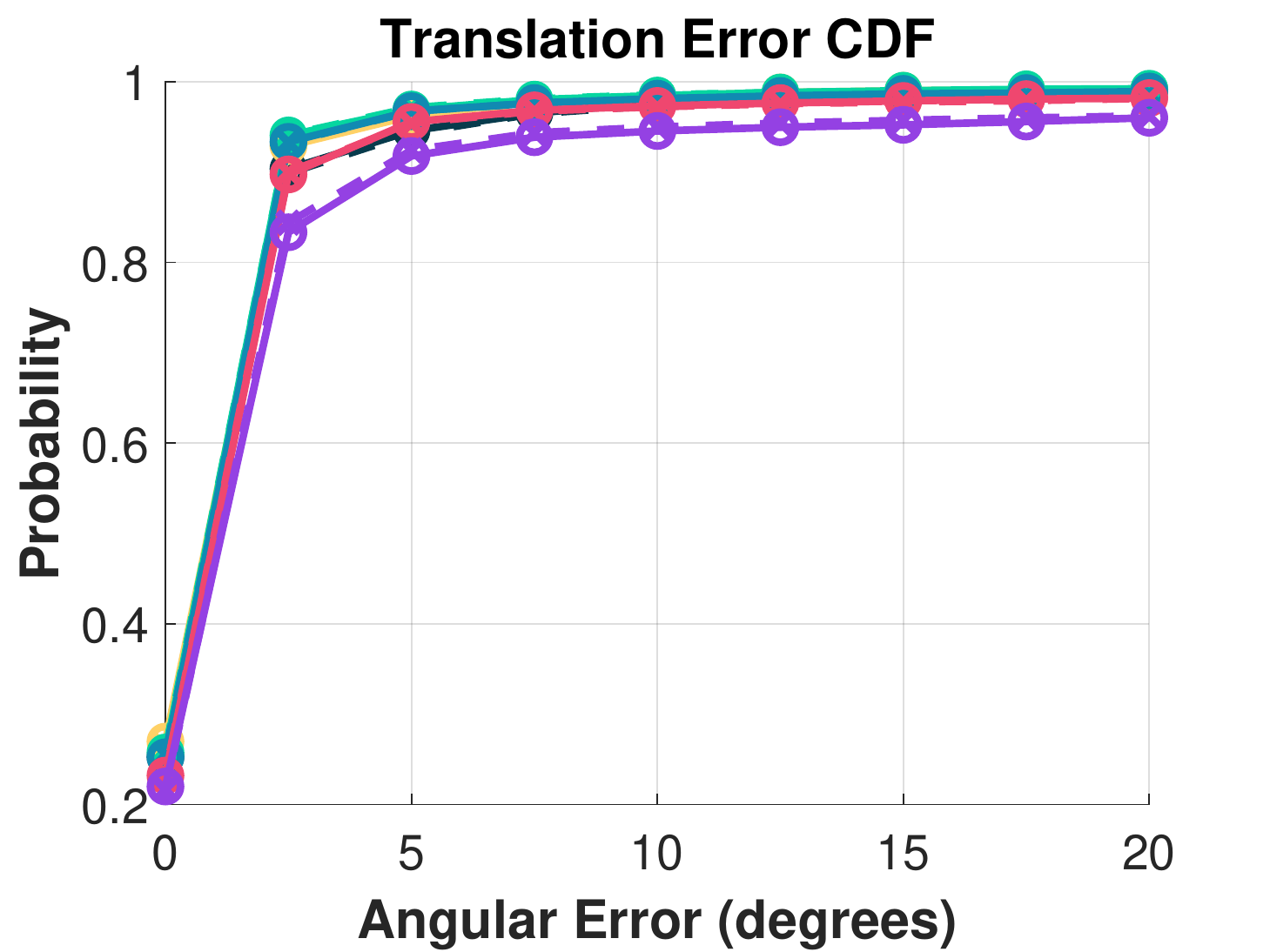}
		\includegraphics[width=0.32\columnwidth,trim={1mm 0mm 10mm 1mm},clip]{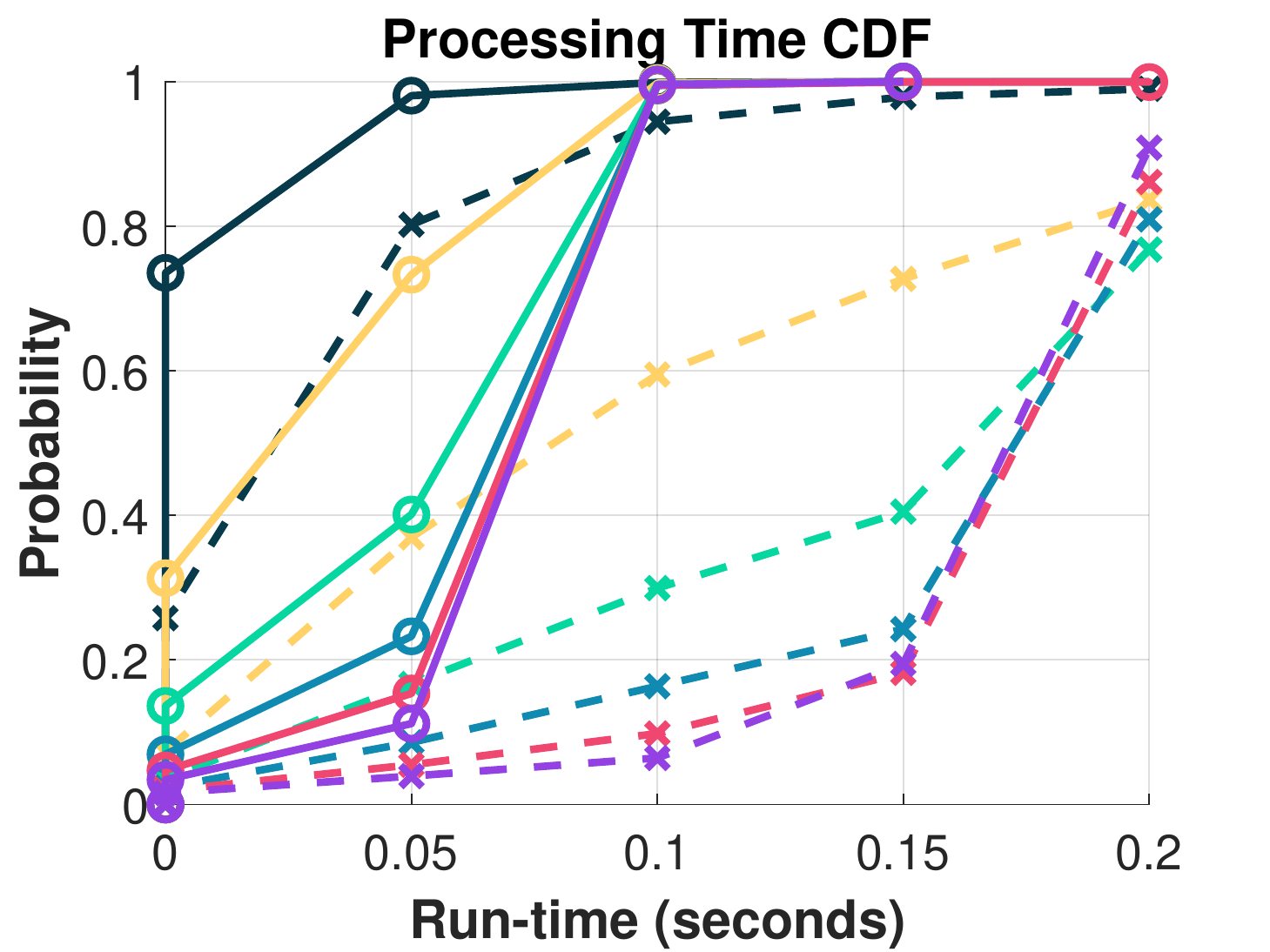}
		\caption{Fundamental matrix estimation}
	\end{subfigure}
	\end{center}
	\caption{\textbf{CDF on KITTI}.
	The cumulative distribution functions (CDF) of the rotation and translation errors (degrees) and run-time (seconds) of epipolar geometry estimation by USAC with and without the proposed neural sample filtering on 47700 image pairs from sequences $6$--$10$ from the KITTI dataset~\cite{geiger2012we}.
	Colors indicate the frame difference, e.g. the red curve uses a frame difference of 5.
	The corresponding mean values are in Table~\ref{table:results_driving}.
	}
	\label{fig:driving_cdf}
\end{figure}

In this section, we compare several variations of our filtering network on the quality of the ordering they induce on minimal samples. 
In practice, for each tested baseline and for each test image, we take a pool of $N = 2^{16}$ random minimal samples, and sort them according to the predicted model score to have the best samples first. 
We then select the first $k$ minimal samples for filtering rates $r \in \{1, 2, 4, 8, 16, 32, 64, 128, 256\}$ where $k = N / r$, and measure its precision, defined as the rate of samples leading to models with less than $10^\circ$ of rotation and angular translation error, and less than 2 pixels of Sampson error.
We test on a motion-constrained autonomous driving dataset and on a weakly-constrained image collection dataset. We use KITTI~\cite{geiger2012we} for the motion-constrained scenario, and use sequences 0 to 4 for training, sequence 5 for validation and early stopping, and sequences 6 to 10 for testing.
We train and validate using random frame differences between 1 and 7, and test with random frame differences between 1 and 5. 
\begin{figure}[t]
	\begin{center}
	\begin{subfigure}[t]{0.93\columnwidth}
		\includegraphics[width=0.32\columnwidth,trim={1mm 0mm 10mm 1mm},clip]{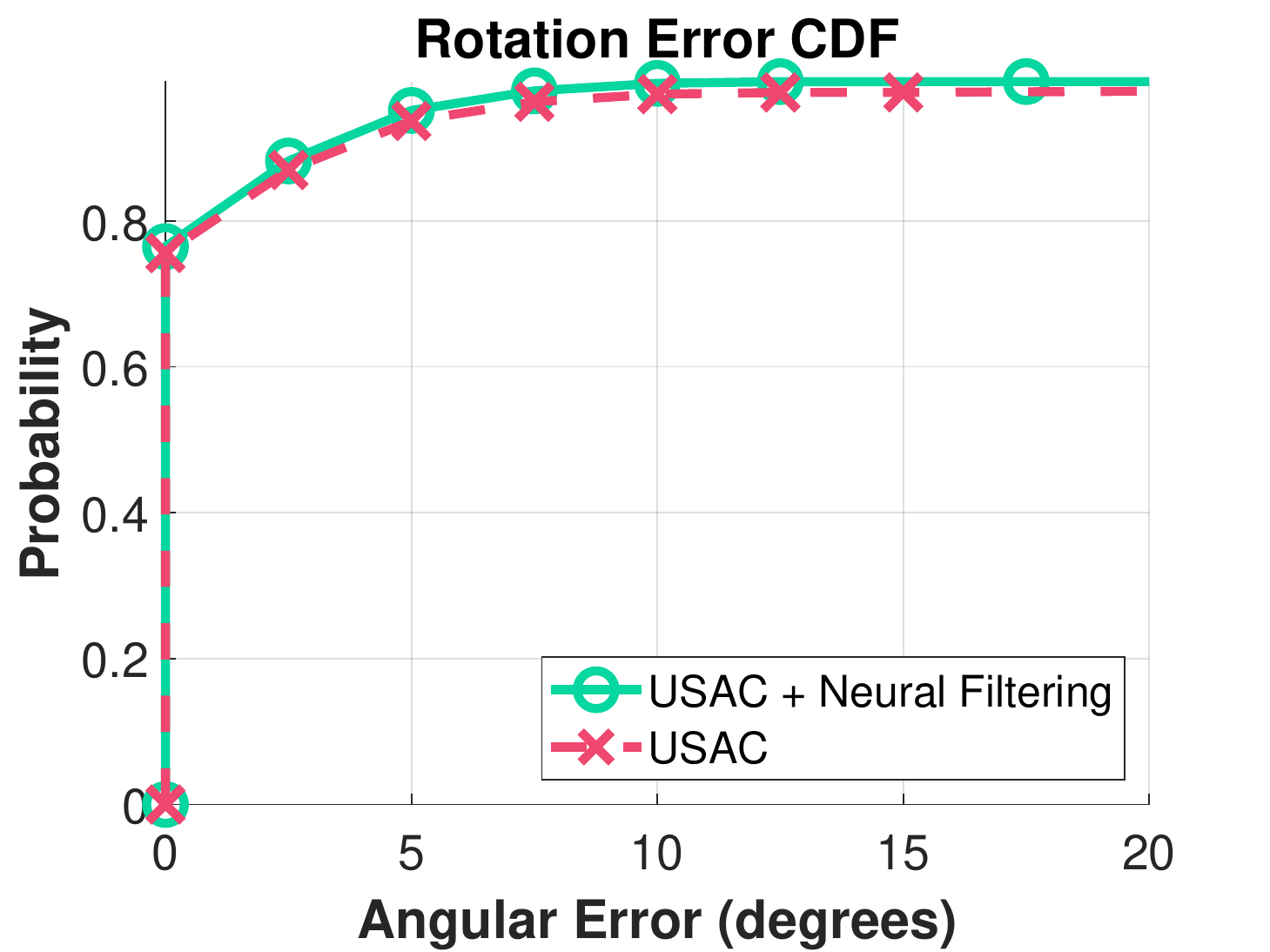}
		\includegraphics[width=0.32\columnwidth,trim={1mm 0mm 10mm 1mm},clip]{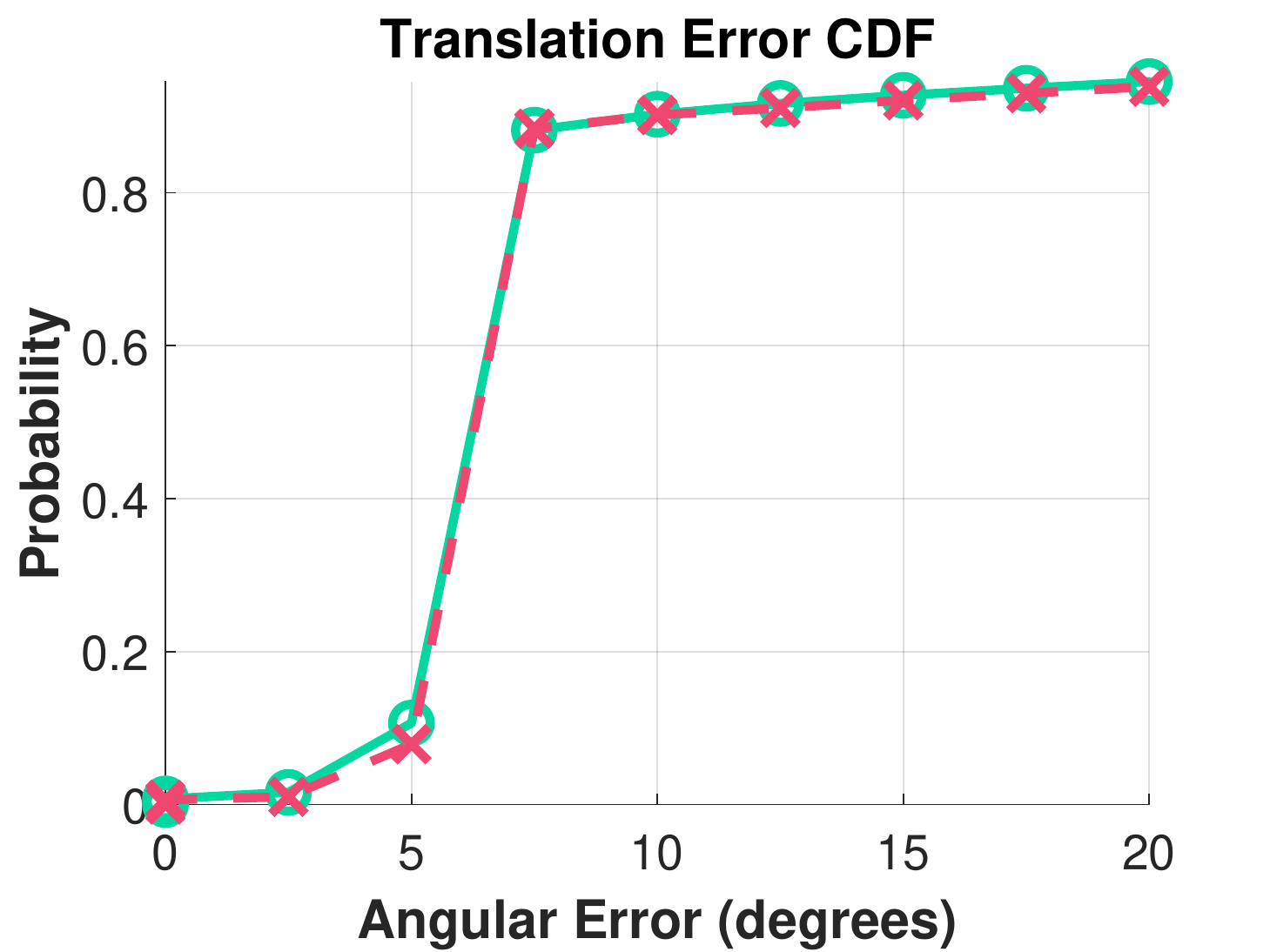}
		\includegraphics[width=0.32\columnwidth,trim={1mm 0mm 10mm 1mm},clip]{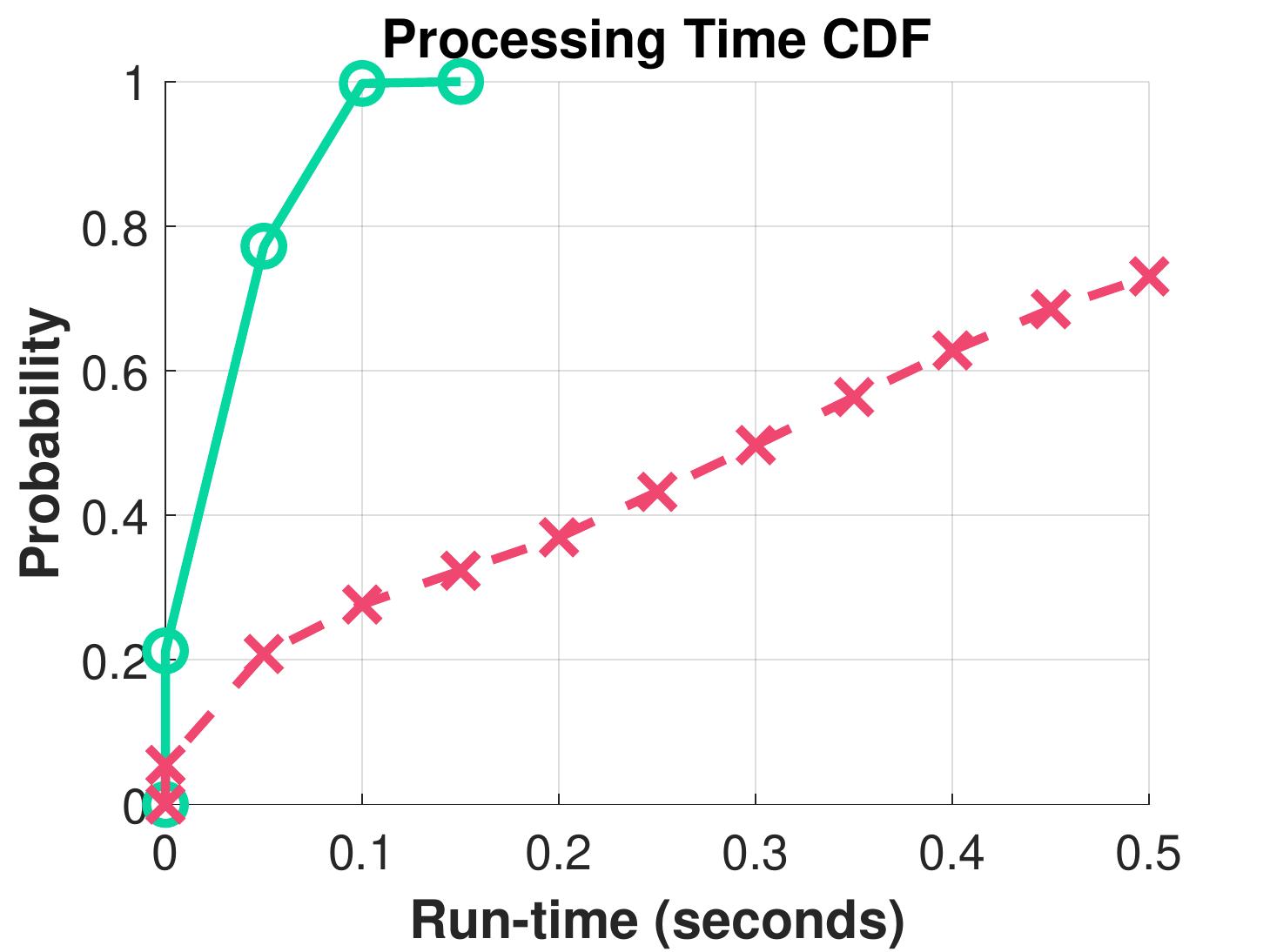}
		\caption{Essential matrix estimation}
	\end{subfigure}
	\begin{subfigure}[t]{0.93\columnwidth}
		\includegraphics[width=0.32\columnwidth,trim={1mm 0mm 10mm 1mm},clip]{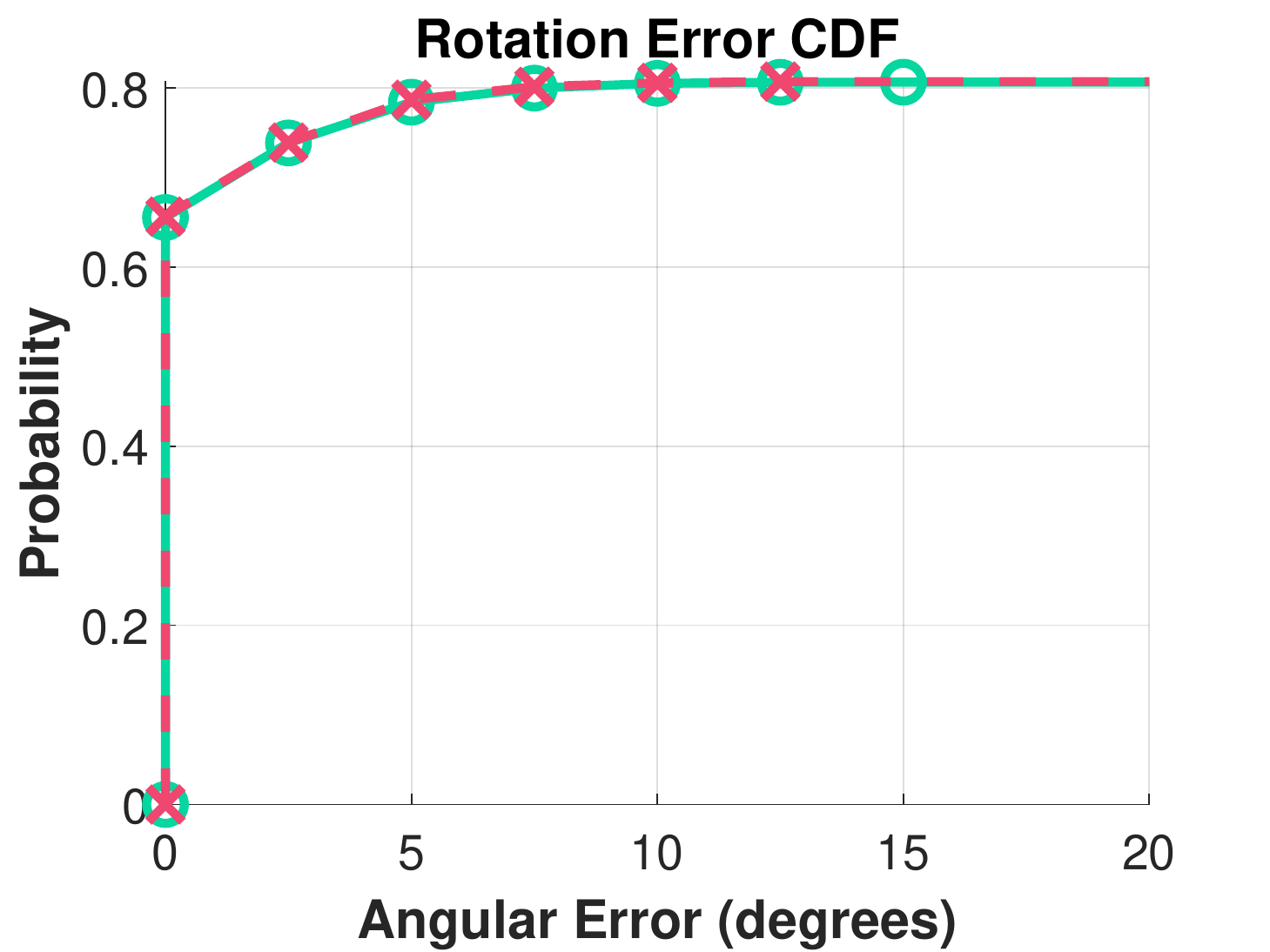}
		\includegraphics[width=0.32\columnwidth,trim={1mm 0mm 10mm 1mm},clip]{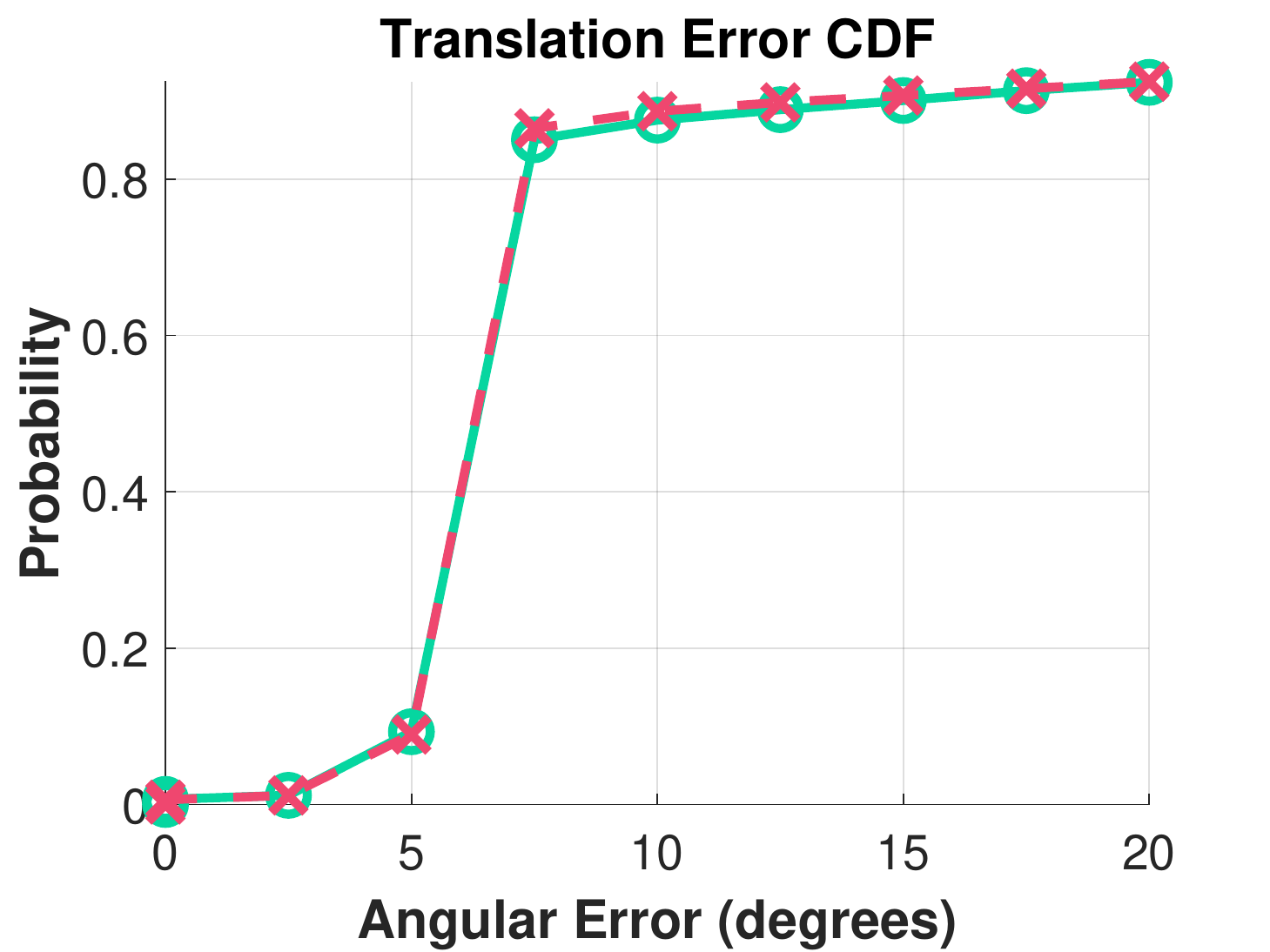}
		\includegraphics[width=0.32\columnwidth,trim={1mm 0mm 10mm 1mm},clip]{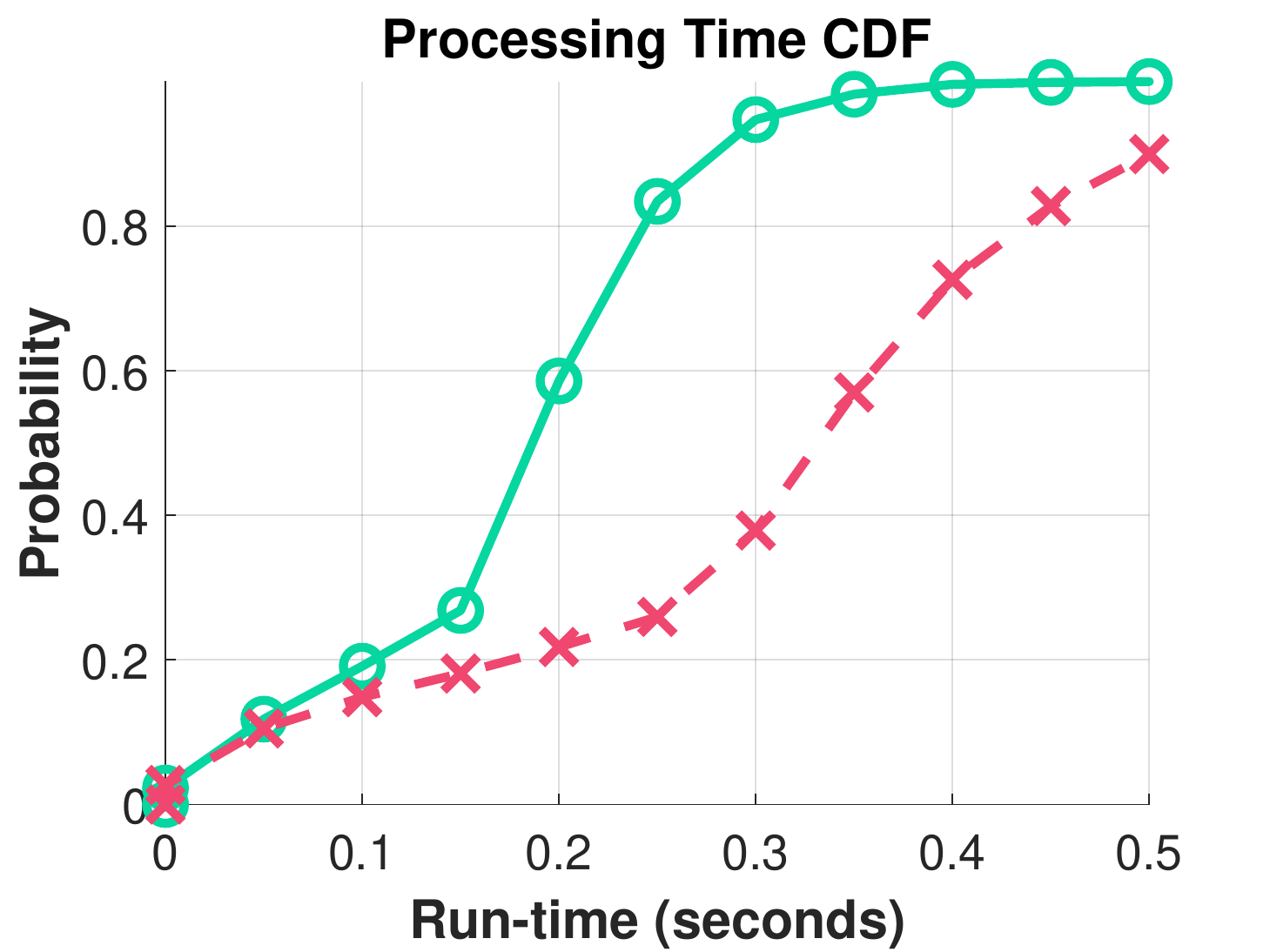}
		\caption{Fundamental matrix estimation}
	\end{subfigure}
	\end{center}
	\caption{\textbf{CDF on Malaga}.
	The cumulative distribution functions (CDF) of the rotation and translation errors (degrees) and run-time (seconds) of epipolar geometry estimation by USAC~\cite{raguram2013usac} with and without the proposed neural minimal sample filtering on 27147 image pairs from the Malaga dataset~\cite{blanco2014malaga}. The corresponding mean values are in Table~\ref{table:results_driving}.
	}
	\label{fig:malaga_cdf}
\end{figure}
In KITTI, we mostly see forward and turning motions with limited speed, thus a strong motion prior can be learned. 
For the weakly-constrained scenario, we use the PhotoTourism~\cite{snavely2006photo} data from the 2020 CVPR RANSAC Tutorial~\cite{cvpr2020ransactutorial}, with the suggested train-validation-test split. 
This dataset does not consist of image sequences but rather of crowd-sourced image collections of famous landmarks, therefore the motion prior that this data can exhibit is very limited. 
However, there are still distinctive motion patterns originating, e.g., from the fact that people usually align their photos with the gravity direction and, also, the range of translations along the vertical axis is limited. 

We compare the following alternatives:
(\romannumeral1)    \textbf{w/o expert branch}: NeFSAC is trained only with branches $B_1$ and $B_2$ with Sampson error and pose error, as described in Section~\ref{sec:supervision}.
(\romannumeral2)    \textbf{with expert branch}: NeFSAC, in addition to $B_1$ and $B_2$, is trained with a further branch $B_3$ with the expert supervision defined in the supplementary material.
(\romannumeral3)    \textbf{w/o branching}: NeFSAC is directly trained to infer the complete label $l_1l_2$ indicating low Sampson error and low pose error. Note that this is the same metric we use for testing, therefore this baseline has an intrinsic advantage in the testing process. Nonetheless, we show that branching leads to superior performance.
(\romannumeral4)    \textbf{domain shift}: NeFSAC has been trained on a significantly different dataset from the one at test. We use the model trained on KITTI for the test on PhotoTourism, and the model trained on PhotoTourism for the test on KITTI. We use no expert branch for this baseline.

In Figure~\ref{fig:phototourism_filter_accuracy}, we show results on 4000 image pairs from the PhotoTourism validation set. Our neural filtering model without any expert branch improves precision (as defined above) by 2.5 times on average at peak filtering rates compared to the original minimal sample pool. Expert knowledge, being inaccurate and hard to formulate in this context, has a slightly detrimental influence. The baseline without any branching, even though it is the only one trained end-to-end to optimize the test metric, does not keep up with the branched alternatives on higher filtering rates. Even with an extreme domain gap, the model trained on KITTI still manages to improve the quality of the original sample pool, showing that NeFSAC is very robust to distribution shifts. We attribute this robustness to the limited information that the neural filtering has at inference time, since our model never observes the global configuration of correspondences.

In Figure~\ref{fig:kitti_filter_accuracy}, we show results on 4000 image pairs from the KITTI sequences 6 to 10. The strong motion statistics on this dataset allow NeFSAC to improve the precision of the minimal sample pool \textbf{by over one order of magnitude} (18x) at peak filtering rates. Filtering is close-to-perfect up until 25\% keep rate. On this dataset, we do not observe significant differences between the three main variants, likely due to the presence of a very simple and discriminative motion model that is well learned by all the baselines trained on KITTI. However, we can observe that the model with expert branch performs the best on this domain, where the expert supervision is more adherent to the real dataset statistics. Interestingly, we found that the learned weight $w_3$ on the expert branch of this model is close to zero: this branch is not playing a significant role in the final prediction of the network, even though it still had a positive impact as a prior in learning good features for the other branches during training. The model trained on PhotoTourism, not taking advantage of the restricted motion statistics of this test set, still improves the precision of the original minimal sample pool by a factor up to 2, very close to its original performance on the PhotoTourism test.

Finally, in Figure~\ref{fig:expert_branch_littledata}, we test the influence of our expert branch in conditions of data scarcity. We keep the same evaluation protocol and test set as in Figure~\ref{fig:kitti_filter_accuracy}, but we train NeFSAC only on 250 image pairs from KITTI, all taken from sequence 0 and fixed frame difference of 4. We also report the baseline NeFSAC trained on the full training set for comparability. We observe that the impact of the expert branch is very significant when little training data is available, halving the gap to the baseline filtering accuracy.
This scenario can be very important when training NeFSAC in a new domain with limited data. 

\begin{table}[t]
\centering
\setlength{\tabcolsep}{4pt}
\resizebox{1.0\columnwidth}{!}{\begin{tabular}{ | l |  c c c c | c c c c | }
 	\hline
 	     & \multicolumn{4}{c|}{KITTI (47700 pairs)} & \multicolumn{4}{c|}{Malaga (27147 pairs)} \\
 	\hline
 		  USAC & $\epsilon_\textbf{R}$ ($^\circ$) & $\epsilon_\textbf{t}$ ($^\circ$) & $t$ (ms) & \# models  & $\epsilon_\textbf{R}$ ($^\circ$) & $\epsilon_\textbf{t}$ ($^\circ$) & $t$ (ms) & \# models \\
 	\hline
 		 {\small w/o NF (\textbf{E})} & \textbf{4.3} & {2.5} & 234.8 & \phantom{1}941 & 3.3 & {8.9} & 350.0 & 3225 \\
 		 {\small w/\phantom{o} NF (\textbf{E})} & \textbf{4.3} & \textbf{2.3} & \phantom{1}\textbf{69.7} & \phantom{1}\textbf{260} & \textbf{1.9} & \textbf{8.7} & \phantom{1}\textbf{34.0} & \phantom{1}\textbf{753} \\
 	\hline
 		 {\small w/o NF (\textbf{F})} & \textbf{4.2} & 2.7 & 213.4 & 1974 & \textbf{1.4} & 9.0 & 380.2 & 3837  \\
 		 {\small w/\phantom{o} NF (\textbf{F})} & \textbf{4.2} & \textbf{2.3} & \phantom{1}\textbf{85.9} & \phantom{1}\textbf{357} & \textbf{1.4} & \textbf{8.6} & \phantom{1}\textbf{77.1} & \textbf{467} \\
 	\hline
\end{tabular} }
\caption{ \textbf{Results on KITTI and Malaga}. The average rotation and translation errors (degrees), the run-time (milliseconds), and the number of models tested inside USAC~\cite{raguram2013usac} on the KITTI~\cite{geiger2012we} and Malaga~\cite{blanco2014malaga} datasets for essential (\textbf{E}) and fundamental matrix (\textbf{F}) estimation. NeFSAC provides great speed-ups while improving accuracy as well. The corresponding CDFs are in Figs.~\ref{fig:driving_cdf},\ref{fig:malaga_cdf}. }
\label{table:results_driving}
\end{table}

\begin{table}[t]
\centering
\setlength{\tabcolsep}{4pt}
\resizebox{1.0\columnwidth}{!}{\begin{tabular}{ | l |  c c c c | c c c c | }
 	\hline
 	     & \multicolumn{4}{c|}{Essential matrix} & \multicolumn{4}{c|}{Fundamental matrix} \\
 	\hline
 		  USAC & $\epsilon_\textbf{R}$ ($^\circ$) & $\epsilon_\textbf{t}$ ($^\circ$) & $t$ (ms) & \# models  & $\epsilon_\textbf{R}$ ($^\circ$) & $\epsilon_\textbf{t}$ ($^\circ$) & $t$ (ms) & \# models \\
 	\hline
 		 {\small w/o NF} & 2.7 & 7.9 & 805.1 & 4550 & 4.8 & 22.5 & 154.6 & 5559  \\
 		 {\small w/\phantom{o} NF} & \textbf{2.1} & \textbf{6.1} & \phantom{1}\textbf{76.5} & \phantom{1}\textbf{364} & \textbf{3.9} & \textbf{17.9} & \phantom{1}{61.7} & \phantom{1}764 \\
 	\hline
 		 {\small w/\phantom{o} NF*} & 2.6 & 7.8 & 103.0 & \phantom{1}660 & 4.8 & 22.3 & \phantom{1}\textbf{61.2}  & \phantom{1}\textbf{740} \\
 	\hline
\end{tabular} }
\caption{ \textbf{Results on PhotoTourism}. The median rotation and translation errors (degrees), the average run-time (milliseconds), and the number of models tested inside USAC~\cite{raguram2013usac} on the PhotoTourism~\cite{snavely2006photo} (from~\cite{cvpr2020ransactutorial}; 52200 image pairs) dataset. NF* is trained on KITTI~\cite{geiger2012we}. NeFSAC provides great speed-ups while improving accuracy as well. The corresponding CDFs are in Fig.~\ref{fig:photo_cdf}. }
\label{table:results}
\end{table}

\subsection{Comparative Experiments within RANSAC}

The interaction between several components in RANSAC is non-trivial: having observed an improvement of ten times in the average precision of a pool of random minimal samples does not necessarily translate into the equivalent speed-up of ten times in any RANSAC. 
In this section, we examine the effect of NeFSAC on a representative RANSAC variant with state-of-the-art components. 
We choose USAC~\cite{raguram2013usac} with cheirality tests, PROSAC sampling~\cite{chum2005matching}, LO-RANSAC~\cite{chum2003locally}, and SPRT~\cite{chum2008optimal} as preemptive verification.
While there might be variants leading to better accuracy, e.g. MAGSAC++~\cite{barath2019magsacpp}, their low run-times still come from SPRT and PROSAC, therefore similar speed-ups are expected.
We show additional experiments in the supplementary material.

\begin{figure}[t]
	\begin{center}
	\begin{subfigure}[t]{0.93\columnwidth}
		\includegraphics[width=0.32\columnwidth,trim={1mm 0mm 10mm 1mm},clip]{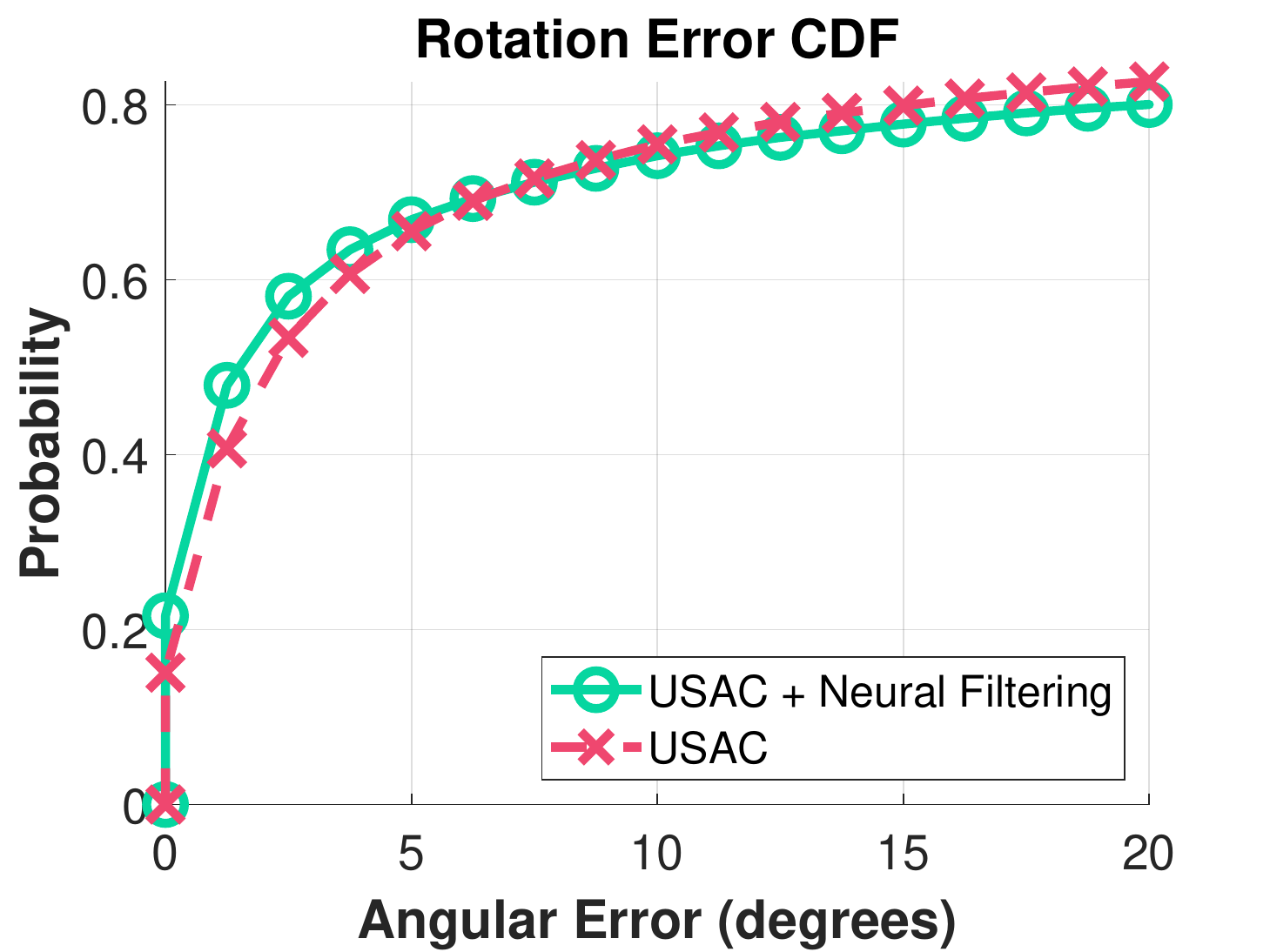}
		\includegraphics[width=0.32\columnwidth,trim={1mm 0mm 10mm 1mm},clip]{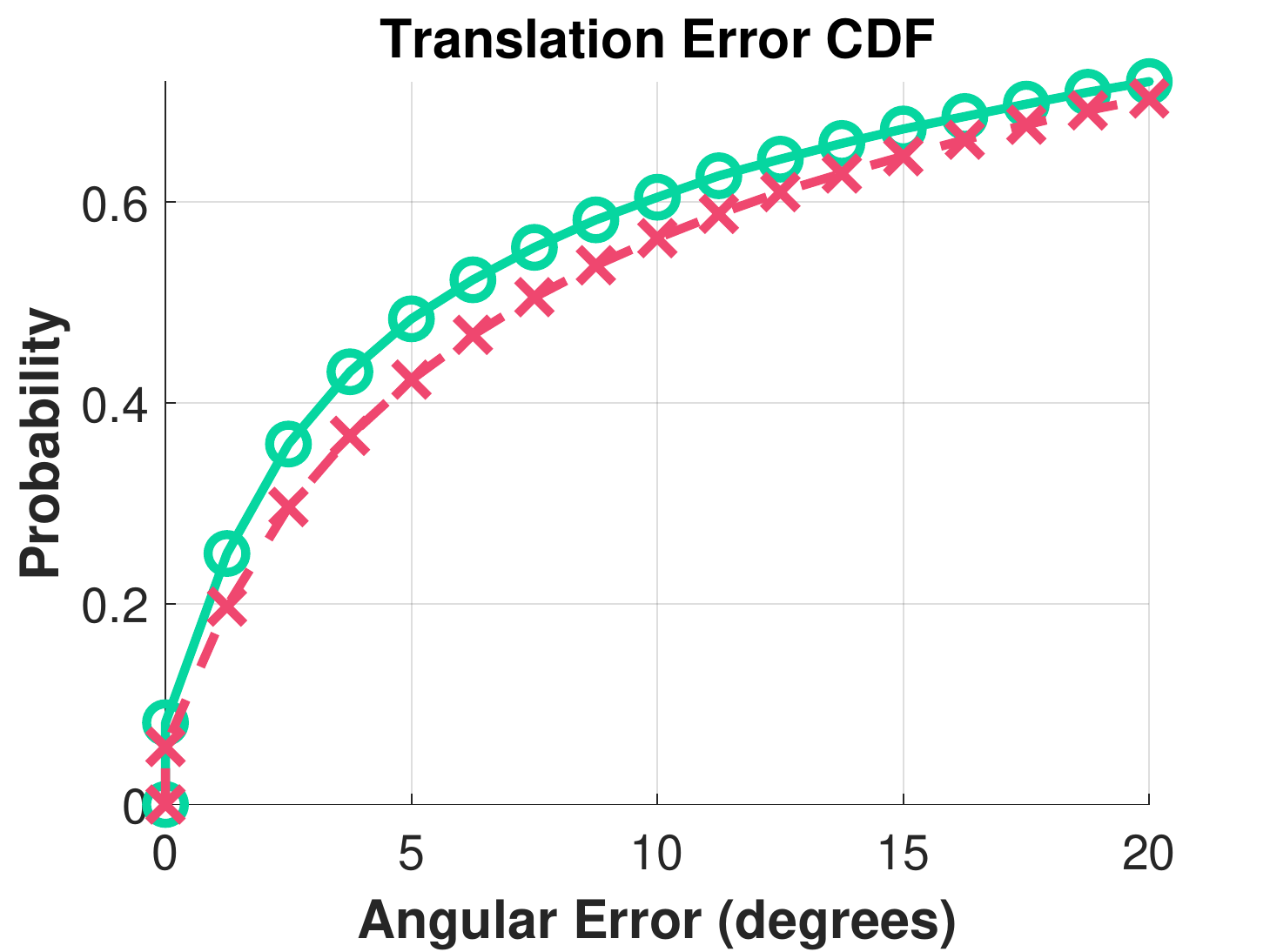}
		\includegraphics[width=0.32\columnwidth,trim={1mm 0mm 10mm 1mm},clip]{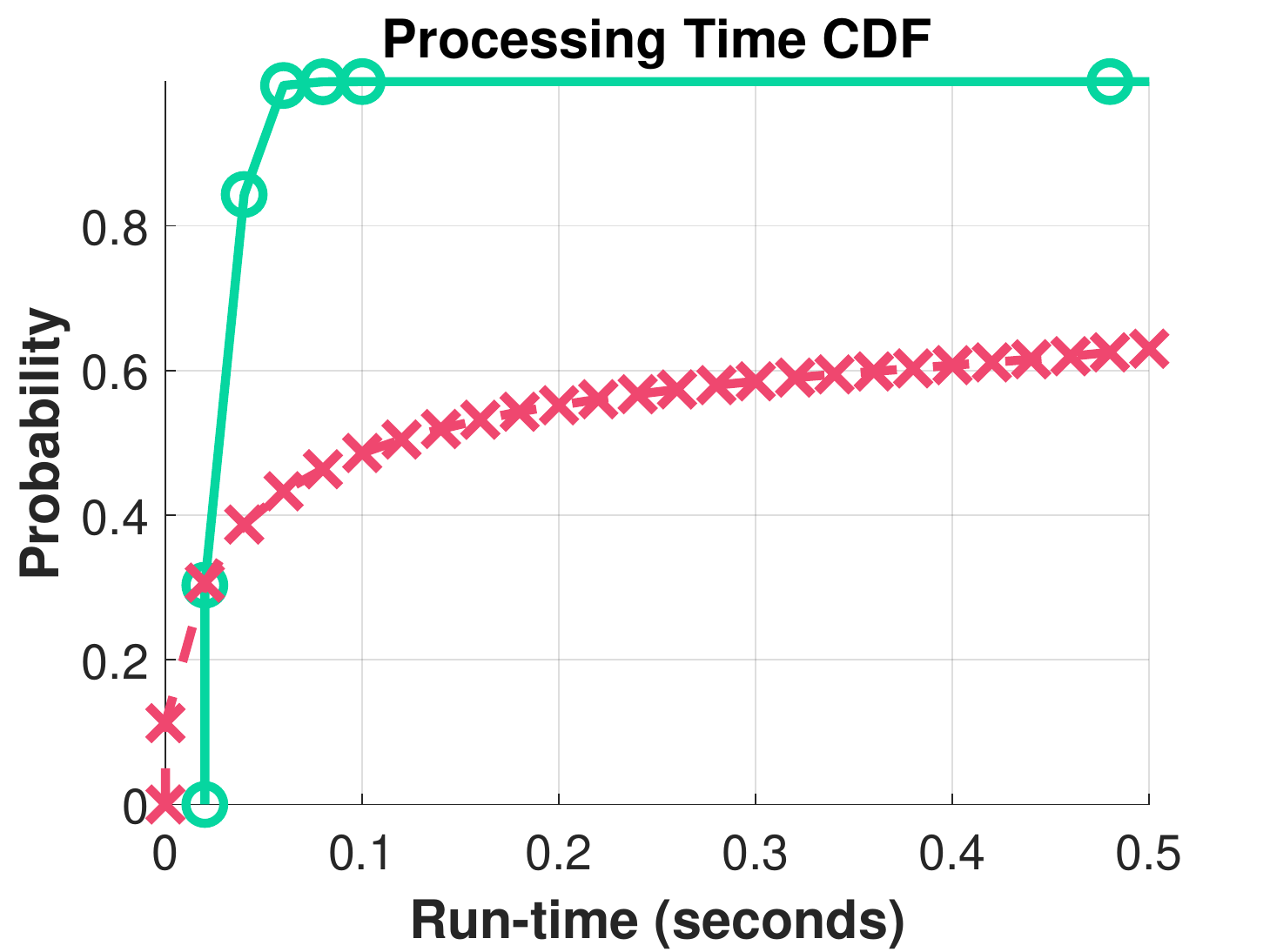}
		\caption{Essential matrix estimation}
	\end{subfigure}
	\begin{subfigure}[t]{0.93\columnwidth}
		\includegraphics[width=0.32\columnwidth,trim={1mm 0mm 10mm 1mm},clip]{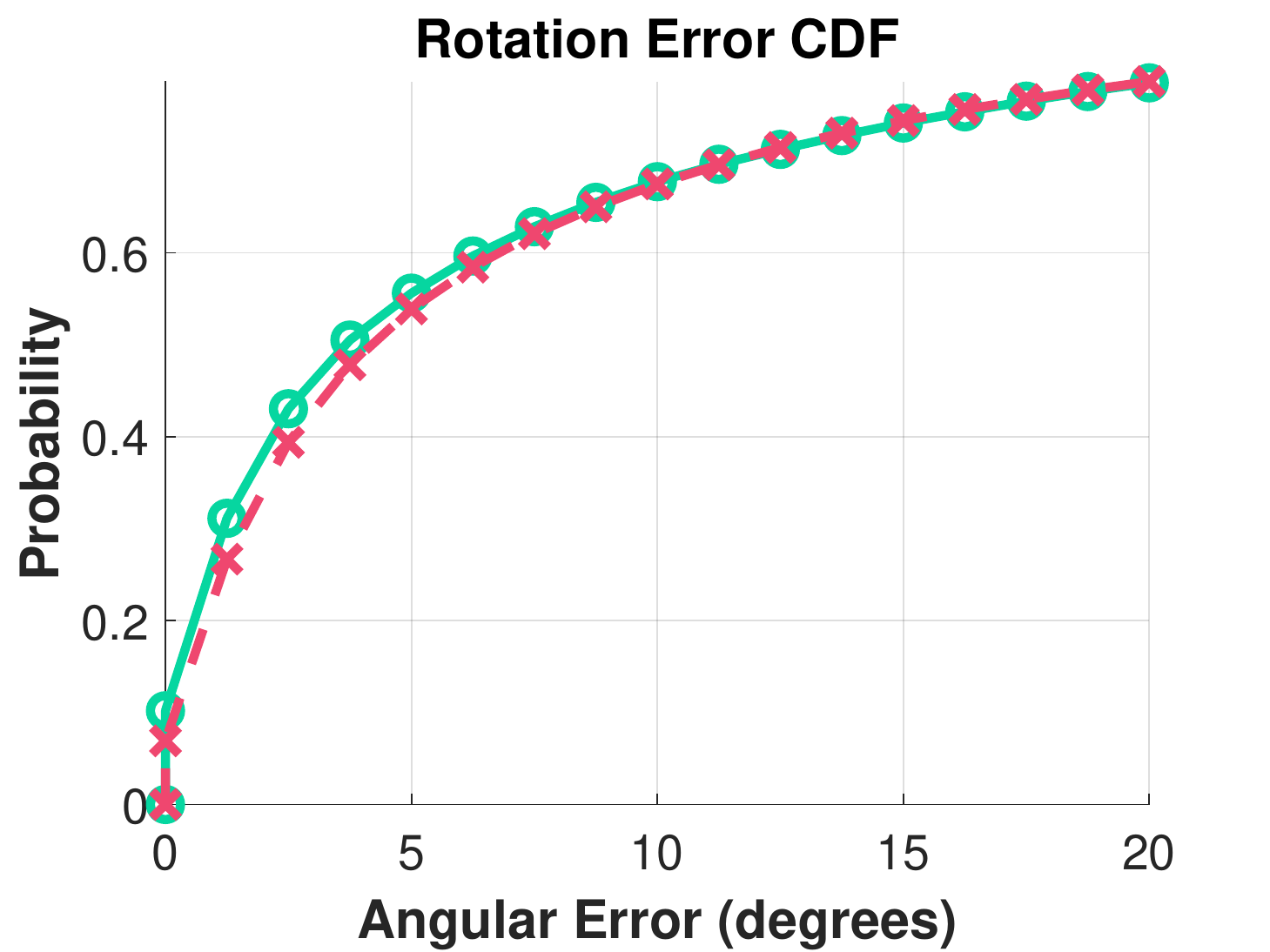}
		\includegraphics[width=0.32\columnwidth,trim={1mm 0mm 10mm 1mm},clip]{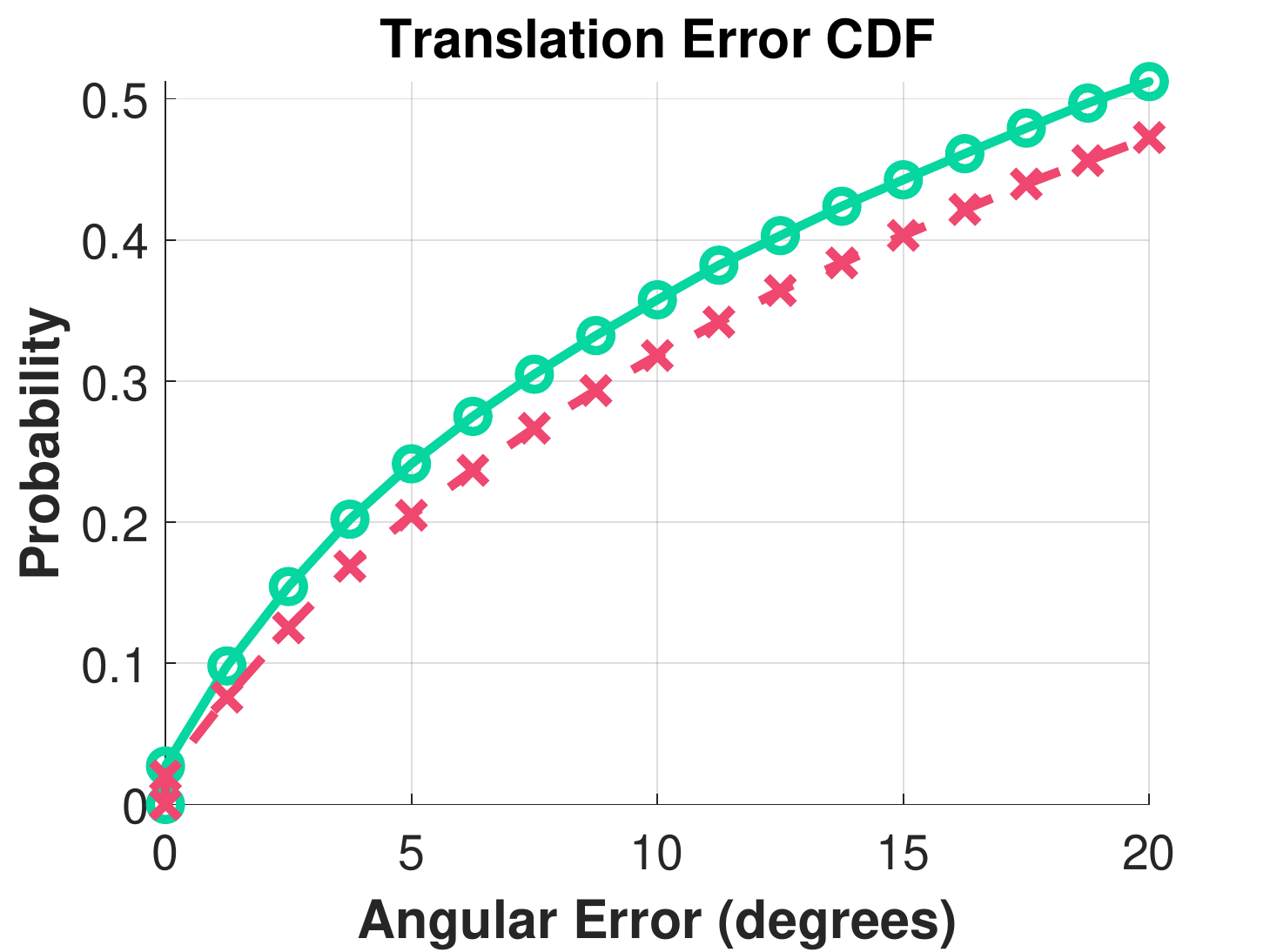}
		\includegraphics[width=0.32\columnwidth,trim={1mm 0mm 10mm 1mm},clip]{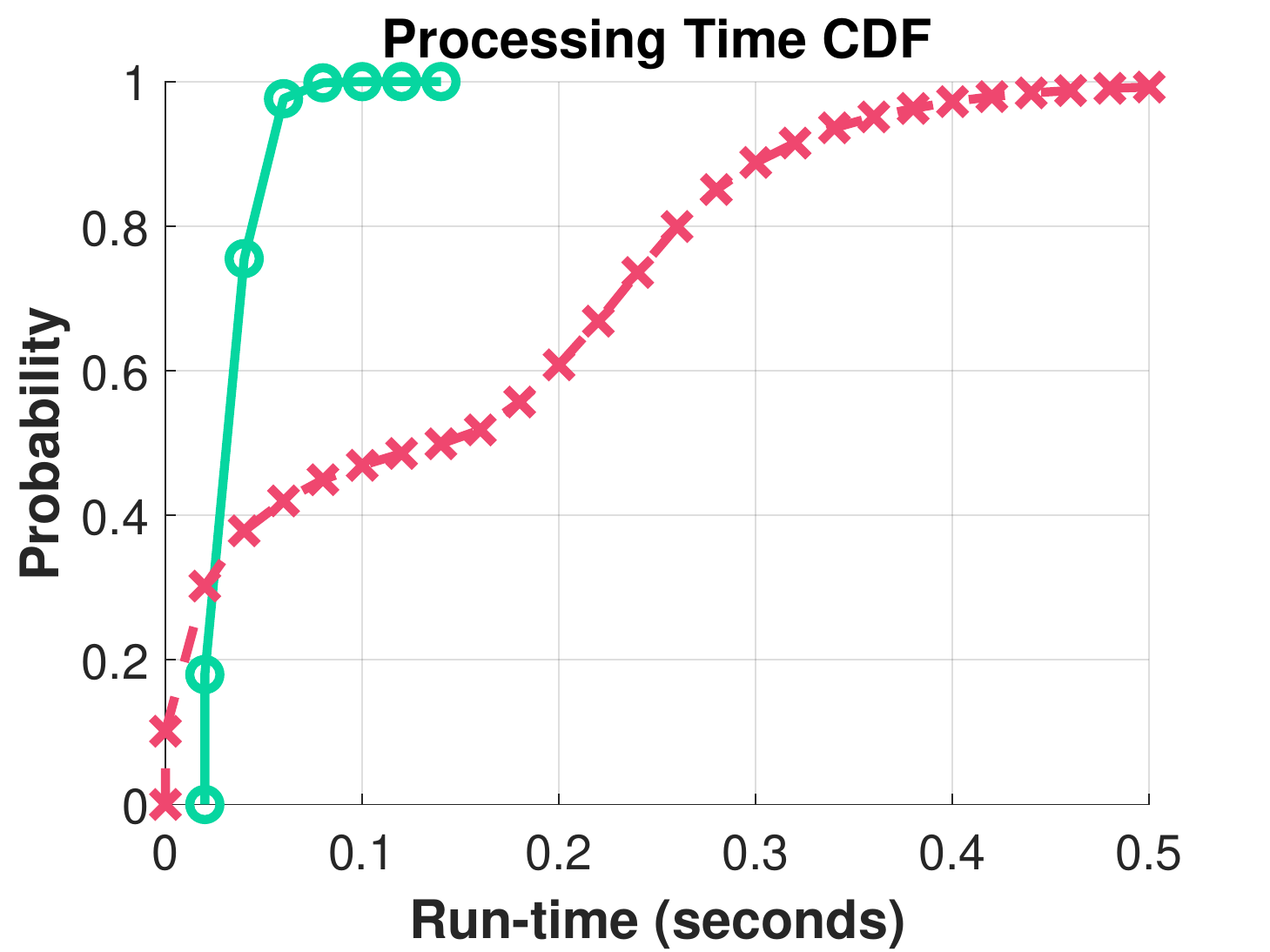}
		\caption{Fundamental matrix estimation}
	\end{subfigure}
	\end{center}
	\caption{\textbf{CDF on PhotoTourism}. The cumulative distribution functions (CDF) of the rotation and translation errors (in degrees) and run-time (in seconds) of epipolar geometry estimation by USAC~\cite{raguram2013usac} with and without the proposed neural minimal sample filtering on a total of 52200 image pairs from the PhotoTourism dataset as used in~\cite{cvpr2020ransactutorial}. 
	The corresponding mean values are in Table~\ref{table:results}. }
	\label{fig:photo_cdf}
\end{figure}

In the following, we compare the rotation error $\epsilon_R$, translation error $\epsilon_t$ and run-time $t$ of USAC with and without NeFSAC filtering for the case of autonomous driving and for the case of unstructured image collections.

\textbf{Autonomous driving.}
We train NeFSAC on KITTI~\cite{geiger2012we} on sequences 0 to 4 with random frame differences between 1 and 7, and use sequence 5 for validation. 
We train separate models for Essential matrix estimation and Fundamental matrix estimation. 
We then test such models on the KITTI sequences 6 to 10 as well as on the Malaga~\cite{blanco2014malaga} dataset to test for generalization. 
We report results in Figures~\ref{fig:driving_cdf} and \ref{fig:malaga_cdf} and Table~\ref{table:results_driving}.
On KITTI, NeFSAC+USAC is over three times faster on E estimation and more than twice as fast on F estimation compared to USAC, with slightly better accuracy.
On Malaga, NeFSAC achieves a \textbf{ten-fold} speed-up and reduces the average rotation error by 1.4 degrees on E estimation, and a five-fold speed-up on F estimation, \textbf{despite being trained on KITTI}.

\textbf{PhotoTourism.}
We train NeFSAC on the PhotoTourism~\cite{snavely2006photo} data provided by the 2020 CVPR RANSAC Tutorial~\cite{cvpr2020ransactutorial}. We use the standard split for training, validation and test. Results are reported in Figure~\ref{fig:photo_cdf} and Table~\ref{table:results}. NeFSAC improves run-time \textbf{again by one order of magnitude} on E estimation and two-fold on F estimation, while providing an \textbf{important improvement of accuracy} on both. Our aggressive filtering setup, tuned for challenging image pairs, causes a small overhead on the easy tail of the distribution, suggesting the use of adaptive strategies. We further perform an extreme generalization test training NeFSAC on KITTI (reported as NF* in Table~\ref{table:results}). While the model trained on PhotoTourism is superior, NeFSAC still manages to bring a very significant speed-up and some improvement in accuracy even under such extreme domain shift. This motivates us to claim that our model is partly learning very general knowledge on the task, and can robustly stay well above the worst-case scenario where it degenerates to the baseline RANSAC.

\section{Conclusions}

In this paper we proposed NeFSAC, a novel framework for Neural Filtering of minimal samples in RANSAC that can be seamlessly integrated into any existing RANSAC pipeline. 
NeFSAC learns to predict the quality of minimal samples by their crude pixel coordinates to filter out the ones consistent with unlikely or impossible motions and common poorly-conditioned configurations. 
We showed that NeFSAC can learn stronger filters when a constrained motion is present in the training data, but can be very discriminative even in datasets without a strong motion prior, like in general image collections, while being very robust to domain shifts. We showed that, in practice, NeFSAC can reduce the run-time by \textbf{one order of magnitude} in modern state-of-the-art RANSAC variants on Essential and Fundamental matrix estimation while often significantly \textit{improving} estimation accuracy.

{ 
\textbf{Acknowledgments:}
This work was supported by the ETH Zurich Postdoctoral Fellowship and the Google Focused Research Award. First published in Proceedings of the 17th European Conference on Computer Vision (ECCV 2022) by Springer Nature. Reproduced with permission from Springer Nature.
}

\section{Supplementary material}

\subsection{Implementation details}
\label{sec:impl_details}

All MLPs in our network use leaky ReLU activation with slope 0.01, except for the final prediction branch which uses the sigmoid activation function. Labels $l_i$ for pose and epipolar errors $e_i$ are assigned the value zero if $e_i > e_\text{max}$, the value of one if $e_i < e_\text{min}$ and are linearly interpolated for error values in $[e_\text{min}, e_\text{max}]$.
We set $e_\text{max} = 5pix$ and $e_\text{min}=2pix$ for epipolar error using the Sampson distance and $e_\text{max} = 30^\circ$ and $e_\text{min} = 5^\circ$ for pose errors using the maximum between translation and rotation errors. When the minimal solver finds multiple solutions, we score the sample according to the best one, and when it fails to find any solution we consider it equivalent to the maximum possible error. The Sampson error assigned to a minimal sample is the maximum Sampson error of all of its correspondences with respect to the ground truth epipolar geometry. All correspondences both for training and for the experimental validation are obtained by matching SIFT~\cite{lowe1999object} with ratio-test filtering of 0.8.

The error of the expert branch for autonomous driving takes inspiration from several existing works on planar constrained motion~\cite{ChoiK18,hajder2019relative,scaramuzza20111}. We assume that the correct motion is a rotation purely around the vertical axis and a translation without any vertical component, and penalize the maximum rotation or translation component that deviates from such assumption. Finally errors are mapped to classification labels with the same thresholds as with the previously defined pose errors.
%
%
%
We further define a tentative error for an expert branch to partially capture the weaker regularities in a scenario of PhotoTourism~\cite{snavely2006photo} image collections. We consider a motion plausible when the relative rotation is either almost horizontal or almost vertical, and penalize the deviation from such model.

Since our technique proposed in Section 3.2 of the main paper does not suffice to produce a balanced dataset, we further address its residual imbalance with class weighting. For each branch, we keep a running count of positive and negative samples, and weigh each sample according to the inverse of the frequency of its class~\cite{zadrozny2003cost}.

\subsection{Experiments with MAGSAC++}

In this section, we provide additional experiments of integrating NeFSAC on a substantially different RANSAC variant than USAC~\cite{raguram2013usac}.
We combined NeFSAC with MAGSAC++~\cite{barath2019magsac,barath2019magsacpp} and ran essential matrix estimation on scene British Museum from the PhotoTourism dataset, KITTI with the frame difference of $4$, and the Malaga dataset.
The AUC@$10$ scores of the maximum rotation and translation errors, the average number of models tested, and the run-time in \textit{ms} are shown in Table~\ref{tab:nefmagsac}.
\bgroup
\setlength{\tabcolsep}{10pt}
\def\arraystretch{1.1}
\begin{table}[h]
\centering
\begin{tabular}{ r | c c | c c | c c }
    \hline   
    & \multicolumn{2}{c | }{ \textbf{ PhotoTourism } } & 
    \multicolumn{2}{c | }{ \textbf{ KITTI } } & 
    \multicolumn{2}{c }{ \textbf{ Malaga } } \\
    \hline   
    NeFSAC & w/o & w/\phantom{o} & w/o & w/\phantom{o} & w/o & w/\phantom{o} \\
    \hline   
    AUC@10 $\uparrow$ & 0.82 & \textbf{0.83} & 0.80 & \textbf{0.82} & 0.82 & \textbf{0.86} \\
    \# models $\downarrow$ & 409 & \textbf{153} & 838 & \textbf{204} & 3888 & \textbf{435} \\
    time (ms) $\downarrow$ & \phantom{1}59 & \textbf{\phantom{1}44} & 415 & \textbf{127} & 1456 & \textbf{122} \\
    \hline   
\end{tabular}
\caption{Augmenting MAGSAC++ with NeFSAC.}
\label{tab:nefmagsac}
\end{table}
\\
NeFSAC increases the accuracy, decreases the number of tested models and, thus, the run-time on all datasets

\subsection{Testing NeFSAC with an inlier oracle}
We argue that the contribution of NeFSAC is orthogonal to outlier filters. In this section, we use NeFSAC in combination to an idea inlier probability predictor on KITTI, which orders the available pool of correspondences according to their residuals with respect to the ground truth essential matrix. Due to using the PROSAC sampler, the probability of finding an all-inlier sample early is extremely high.
We report results of USAC + Oracle, with or without NeFSAC, in Table~\ref{tab:neforacle}.
NeFSAC is still able to improve accuracy even in this ideal case, showing that learning to avoid degeneracies can still contribute on top of an ideal inlier selection.

\begin{table}[h]
\centering
\begin{tabular}{ r | c | c }
    \hline   
    & USAC + Oracle & USAC + Oracle + NeFSAC \\
    \hline   
    AUC@10 $\uparrow$ & 0.93 & \textbf{0.95}\\
    \# models $\downarrow$ & 65 & \textbf{47}\\
    time (ms) $\downarrow$ & 9 & \textbf{10}\\
    \hline   
\end{tabular}
\caption{NeFSAC influence with Oracle inlier probability predictor.}
\label{tab:neforacle}
\end{table}

\clearpage
%
%
\bibliographystyle{splncs04}
\bibliography{egbib}
\end{document}